\documentclass[journal]{IEEEtran}
\ifCLASSINFOpdf
\else
\fi
\usepackage{url}
\usepackage{balance}

\usepackage{lipsum}
\usepackage{siunitx}
\usepackage{amssymb}
\usepackage{amsmath} 
\usepackage{hyperref}
\usepackage{cite}
\usepackage{tikz}
\usetikzlibrary{calc,patterns,decorations.pathmorphing,decorations.markings,positioning,backgrounds,arrows.meta,shapes,fit,matrix,spy,shapes.geometric}
\usepackage{booktabs}
\usepackage{import}
\usepackage{float}
\usepackage{adjustbox}

\usepackage{fancyhdr}

\IEEEoverridecommandlockouts
\IEEEpubid{\makebox[\columnwidth]{978-1-5386-5541-2/25/\$31.00~\copyright2025 IEEE \hfill}
\hspace{\columnsep}\makebox[\columnwidth]{ }}

\begin{document}

\title{Force-Based Viscosity and Elasticity Measurements for Material Biomechanical Characterisation with a Collaborative Robotic Arm}

\author{Luca Beber$^{1,2}$, Edoardo Lamon$^{1}$, Giacomo Moretti$^{3}$, Matteo Saveriano$^{3}$,\\ Luca Fambri$^{3}$, Luigi Palopoli$^{1}$, and Daniele Fontanelli$^{3}$
\thanks{$^{1}$Department of Information Engineering and Computer Science, Universit\`a di Trento, Trento, Italy. \tt\small luca.beber@unitn.it}
\thanks{$^{2}$DRIM, Ph.D. of national interest in Robotics and Intelligent Machines.}
\thanks{$^{3}$Department of Industrial Engineering, Universit\`a di Trento, Trento, Italy.}
% \thanks{This research has been funded by the MUR ``Departments of Excellence 2023-27'' program (L.232/2016), by the PNRR project FAIR - Future AI Research (PE00000013), and by the European Union projects INVERSE (GA no. 101136067) and MAGICIAN (GA no. 101120731).}
}

\maketitle
\IEEEpubidadjcol
% \thispagestyle{firstpage}
\vspace{-1.5em}
\begin{center}
\footnotesize
This is the accepted version of the paper published as: \\
L. Beber et al., "Force-Based Viscosity and Elasticity Measurements for Material Biomechanical Characterization With a Collaborative Robotic Arm," \\
\textit{IEEE Transactions on Instrumentation and Measurement}, vol. 74, pp. 1--14, 2025, Art. no. 4013314.\\
doi: \texttt{10.1109/TIM.2025.3581663}. Available at: \url{https://ieeexplore.ieee.org/document/11045727}
\end{center}
\vspace{0.5em}
\begin{abstract}
Diagnostic activities, such as ultrasound scans and palpation, are relatively low-cost. They play a crucial role in the early detection of health problems and in assessing their progression. However, they are also error-prone activities, which require highly skilled medical staff. The use of robotic solutions can be key to decreasing the inherent subjectivity of the results and reducing the waiting list.
For a robot to perform palpation or ultrasound scans, it must
effectively manage physical interactions with the human body, which
greatly benefits from precise estimation of the patient's tissue
biomechanical properties. This paper assesses the accuracy and
precision of a robotic system in estimating the viscoelastic
parameters of various materials, including some tests on ex vivo tissues as a preliminary proof-of-concept demonstration of the method's applicability to biological samples. The measurements are compared against a ground truth derived from silicone specimens with different viscoelastic properties, characterised using a high-precision instrument. Experimental results show that the robotic system's
accuracy closely matches the ground truth, increasing confidence in
the potential use of robots for such clinical applications.

\end{abstract}


% \fi
\IEEEpeerreviewmaketitle

\section{Introduction}
\IEEEPARstart{U}{nderstanding} the mechanical properties of biological materials is crucial for a wide range of medical applications. For example, in prosthetics and tissue engineering, synthetic implants or repairs must replicate the specific characteristics of native tissues.  In robotics, any medical procedure involving direct interaction between a robot and the human body can benefit from knowledge of the tissue’s mechanical properties in the operative field.  This information can improve the control accuracy during a surgical cut, help produce high-quality haptic feedback for teleoperated procedures~\cite{Al-Mouhamed2009DesignTelerobotics}, and enable a fine grained localisation of the tool on or inside the body (e.g., to identify the intercostal regions on the rib cage~\cite{BeberLNFSP24icra}). In robot-based diagnosis, tissue
characteristics can be used as indicators of various diseases. For example, a localized region with increased elasticity in the abdominal area may indicate the presence of a neoplastic lesion. This observation is supported by the findings of~\cite{Greenleaf2003SelectedTissues} on selected
tissues. Similar examinations can be used to assess tissue
ageing~\cite{Lau2008IndentationCartilage} or to detect systemic sclerosis (SSC)~\cite{Dobrev1999InSclerosis}.

Among the mechanical properties of human tissues,
elasticity and viscosity are of particular significance. Elasticity refers to the material's ability to deform under
stress and subsequently return to its original
shape once the stress is removed. Viscosity measures the material's resistance to flow or deformation under continuous
stress. Different tissues and organs exhibit a unique viscoelastic
behaviour due to the combination of elastic and viscous responses;
therefore, accurately characterising these parameters is crucial for accurate diagnosis.

Robotic systems offer clear advantages for analyzing mechanical properties.
Unlike other mechanical devices, a robot can utilize its
kinematics to explore large areas of the body (just as a human physician would
do). 
For example palpations for medical purposes are often performed in anatomically challenging areas, such as the
armpit for lymph node examinations, where standard testing devices are impractical. A robotic arm
provides the necessary degrees of freedom to navigate these complex regions, ensuring consistent and
repeatable assessments, and, thanks to the flexibility of the device, can also be used for other medical
practices, making it also economically convenient for specialised instrumentation capable only of
performing elasticity measurements.
Implementing precise viscoelastic contact models allows robots to determine the contact area, interaction forces, and penetration depth of the end-effector into the tissue or body~\cite{BeberLNFSP24icra}.
However, applying robotic diagnosis in real-time presents challenges: inaccurate estimation of viscoelastic parameters can lead to significant diagnostic errors, patient discomfort, and, in the worst case, injury.
This leads to the central research question addressed in this paper: how accurately can a robotic system equipped with standard force sensors estimate the viscoelastic parameters of human tissue?

\paragraph*{Viscoelastic properties estimation} Several techniques have been developed
for the characterisation of the viscoelastic properties of human tissue. For instance, magnetic resonance imaging (MRI) is used to assess
the viscoelastic properties of biological tissues by measuring shear wave velocity~\cite{Manduca2021MRTerminology,Paul2024Surface-Wave-InducedProperties,Pereira1991DynamicSkin}. Magnetic resonance elastography (MRE) is a powerful, non-invasive technique but is constrained by
expensive and complex instrumentation that requires specialised personnel, making it unsuitable for
large-scale initial screening applications. In
contrast, ultrasound elastography
estimate the viscoelastic properties of a material by analysing the
propagation of ultrasound
waves~\cite{Frulio2013UltrasoundLiver,Li2017MechanicsElastography,Tang2015UltrasoundTechniques,Kumar2010MeasurementApplications,Eder2007PerformanceResolution}. Ultrasound elastography is promising but suffers from operator dependence and is primarily useful for
comparative rather than absolute stiffness measurements. Moreover, it requires expensive ultrasound
equipment. Cretu
et al.~\cite{Cretu2008NeuralApplications} proposed a method to estimate the elastic modulus by measuring
the deformation of the soft body with a laser and then using a neural
network-based estimator. Traditional dynamic testing methods, such as rheometry and dynamic mechanical analysis (DMA),
provide high-precision viscoelastic measurements by applying oscillatory forces at controlled frequencies. However, these techniques require specialised laboratory equipment and are limited to small,
isolated samples, making them impractical for in situ or large-scale applications. In contrast, the
proposed robotic approach enables real-time viscoelasticity estimation over larger surface areas while
reducing operator dependency, bridging the gap between high-precision laboratory techniques and
deployable clinical solutions. Furthermore, as demonstrated in our prior work~\cite{BeberLNFSP24icra},
this approach is also suitable for robotic telehealth applications, where knowledge of the mechanical
properties of contacted tissue can substantially enhance operational safety.
%% , where, for
%% example, one might also be interested in assessing the geometric
%% properties of the tissue under mechanical stimuli.

\paragraph*{Force response models}  Viscoelastic parameters
estimation of a material by means of contact forces relies on
appropriate models for the force response.  The literature offers
simple linear models such as the Kelvin-Voigt or the Maxell
models~\cite{Flugge1975Viscoelasticity}. These models allow for a fast
parameter estimation, but are also known for their limited accuracy,
especially in the case of contacts with small
penetration~\cite{Pappalardo2016Hunt-CrossleySurgery}.  The
Hunt-Crossley model is a potentially better
alternative~\cite{Zhu2021ExtendedModel,haddadi2012real,
Pappalardo2016Hunt-CrossleySurgery} for its ability to encode the
nonlinear behaviour of the forces resulting from the three-dimensional
nature of the contact. The downside is that the model requires a
complex
%and time-consuming offline
procedure to identify the model parameters. The fact that these
parameters do not have direct physical meaning adds much to the
complexity of the problem (viscosity and elasticity cannot be
expressed as a direct function of the \textit{stiffness} and
\textit{damping} of the model). To simplify the parameter estimation,
these models are often applied in conjunction with different
estimation techniques, including nonlinear least squares
regression~\cite{diolaiti2005contact,Schindeler2018OnlineLinearization}
and Kalman
filtering~\cite{Pappalardo2016Hunt-CrossleySurgery,Zhu2021ExtendedModel,Roveda2022SensorlessEstimation,Zhu2023IterativeIdentification}. This
is done assuming a known penetration
depth~\cite{Zhu2021ExtendedModel}, a condition that is impractical in
real-world exams like palpation.

\begin{figure}[t]
  \centering
  %\adjustbox{width=0.75\columnwidth}{
  \input{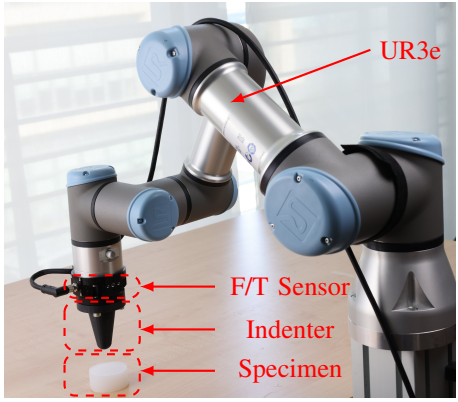} %}
  % \vspace{-0.8cm}
  \caption{Experimental setup consisting of a 6-DoF Ur3e robotic arm, a 6-axis F/T sensor, a 3D printed indenter, and the specimen that is being tested.}
  \label{fig:robot_photo}
\end{figure}

\paragraph*{Paper contribution} This paper presents a method for estimating the local and global viscoelastic properties of soft
materials using a robotic system. The approach enables accurate
reconstruction of contact forces and can be applied both offline
and online. Estimating viscoelastic parameters using non-rigid contact models while accounting for measurement uncertainties requires a physical system capable of applying controlled perturbations to the soft body. As shown in~\autoref{fig:robot_photo}, our choice
fell on a robotic arm equipped with an indenter (as end effector) and
with a precise 6-axis force/torque sensor, that is mounted between the
robot flange and the indenter.
Since it is known in the literature that under static or quasi-static
conditions it is possible to estimate only the elasticity
value~\cite{Hayes1972ACartilage,Sakamoto1996ATest,
Waters1965TheIndentors, Dimitriadis2002DeterminationMicroscope}, 
viscosity and penetration are usually assumed known, because their
estimate requires a much more complex analysis under dynamic
conditions.
%Applications of such a method could span various settings, such as
%medical and industrial environments, which envisions the proximity or
%contact of a robot with the human body, or, in general, with soft
%materials, to increase the awareness and safety of robot controllers
%in physical interactions.
To overcome this limitation, we exploited the \textit{Dimensionality
Reduction (DR)}~\cite{Popov2015MethodFriction} method, commonly
employed in tribology. This method allows us to reduce to a 1D dynamic
equation the 3D contact between an axial-symmetric indenter and a
soft, flat surface.  The equation establishes a relation between the
elasticity and viscosity modulus, and the penetration inside the
material. Since the method exploits the geometry of the elements in
contact, it can be applied to generic indenter shapes, which are, in
our case, the robot end effector tips.  This allows us to lift an
important restriction of previous approaches: the use of spherical
indenter tips~\cite{johnson1985contact} imposed by the Hertzian
theory. The paper presents the main ideas underlying the described
process and the results of the characterisation of the measurement
system using different classes of materials with known model
parameters.

% Whereas in a previous work foams were used~\cite{BeberLPFSF24i2mtc}, in this study we test our method with silicones with different viscolasticity, as they feature dynamic responses under stress comparable to biological tissues~\cite{wells2011medical}. Due to its similarity and compatibility with the human body, silicone is used in prosthetics and in medical devices such as implants and catheters. Moreover, we evaluated the effectiveness of the force reconstructed with the model reduced with the DR by varying the indenter tip shape (flat, spherical, and quartic), and radius.

A subset of the mentioned contributions have been presented in a
conference paper~\cite{BeberLPFSF24i2mtc}, where an elastic
characterisation of testing foams has been proposed to validate the
robot arm as an instrument. In this paper, we extend the method and
the results of~\cite{BeberLPFSF24i2mtc} in several respects: 1) we
include the viscosity in the DR, 2) we propose an online procedure to
perform not only local point-wise measurements but also continuous
measurements on wider surfaces, which present different local
characteristics, 3) we repeat the evaluation of the indenter tip with
silicones and extend it to quartic indenters, which were not
considered in our previous conference paper. An important practical and
theoretical implication of these improvements is in the paper
experimental section, where we analyse both silicones with homogeneous
viscoelasticity and silicones containing harder material inside, which
mimics the presence of stiffer tissues embodied in softer parts.
In~\cite{beber2024robotisedpalpationcancerdetection} we exploited the
online point-wise algorithm to estimate the penetration inside the
soft tissue and the parameters to reconstruct the contact force,
enabling the detection of foreign masses in soft materials. In this
work, we use those results to compare the online estimation of
elasticity with a static estimation and expand the algorithm to have a
continuous estimation of the viscoelasticity parameters. Finally, a preliminary proof-of-concept experiment was conducted to validate the proposed method's ability to estimate the viscoelastic properties of an ex vivo tissue. Measurements were performed both on healthy tissue and on tissue containing a stiffer intrusion.

% \dan{Add the novelties w.r.t. to I2MTC and w.r.t. the literature described previously.}.
% To be cited: \cite{Al-Mouhamed2009DesignTelerobotics,Pereira1991DynamicSkin, Boiko2010MeasurementProperties}
% \cite{Hou2005InstrumentRheometry,Lakes2004ViscoelasticTechniques}
% \cite{Kumar2010MeasurementApplications,Liu2011ASimulation,Xu2011ASimulation,Paul2024Surface-Wave-InducedProperties,Cretu2008NeuralApplications}

The findings in this manuscript, which confirm and extend
the ones on foams of the previous work, show that the level of
accuracy in viscoelasticity estimation and in force reconstruction
reached by the solution is promising for biomedical applications in the near future.

% In addition, the position of the end
% effector and the contact forces were acquired to estimate the
% elasticity coefficient, taking into consideration the shape of the
% end-effector probing tip. At the same time, we determine the position
% of the surface that minimises the residuals of a least squares sum,
% which optimally calculates the elasticity value. 

%-%
%-%

% In~\cite{BeberLNFSP24icra} we have presented \dan{Describe the ICRA24 paper.} è stato citato nell'intro ma non è molto legato ai metodi che usiamo qui, quindi non mi dilungherei a spiegarlo in dettaglio.

% The paper is organised as follows. \dan{Please add the description} Se abbiamo spazio aggiugeremo questo paragrafo.. Da revisore?lettore quando trovo questo paragrafo solitamente lo salto a piè pari

\section{Contact Model with the Dimensionality Reduction Model}
\label{sec:models}
This section presents the fundamental concepts behind the
dimensionality reduction method (DR), proposed by Popov et
al.~\cite{Popov2015MethodFriction} and describing the contact
behaviour between an axial symmetric indenter and a soft material. The
formulation for the elastic contributions was presented in our previous
work~\cite{BeberLPFSF24i2mtc}, while in this study we include
the viscosity component in the contact model. The following section shows how is possible to approximate the response of soft material modelling it as a series of spring-damper elements.

\subsection{Elastic Force Model}
The main idea of the DR method is that the 3D contact between an
indenter of an axially symmetric arbitrary shape and the surface of an
elastic object can be modelled by resorting to a one-dimensional
equivalent model, in which the elastic body is described through a
one-dimensional linearly elastic foundation, consisting of a set of
identical springs positioned at a small distance from each
other~\cite{Popov2015MethodFriction}. The resulting contact stiffness
of each spring is a function of the elastic modulus of the material
and of the distance between neighbour springs, i.e.,
\begin{equation}
  \Delta k_z = E^* \Delta x ,
  \label{eq:dr2}
\end{equation} 
with $\Delta k_z$ being the stiffness of a single spring, $\Delta x$
the spacing between the springs and $E^*$ the so-called effective elastic modulus.

An effective elastic modulus $E^*$ for the substrate can be extracted
knowing the Poisson ratio $\nu$ and the material elastic modulus $E_f$:
\begin{equation}
    E^* = \dfrac{E_f}{1-\nu^2},
    \label{eq:elast}
\end{equation}
which stems from the assumption that the material is in plane-stress in one of the directions perpendicular to the indentation direction, and in plane-strain in the the third direction. 
The force exerted by each spring is expressed as a function of the
deformation using~\eqref{eq:dr2} and \eqref{eq:elast}, that is
\begin{equation}
    \Delta f_{N,i} = \Delta k_z\, u_{z,i} = \frac{E_f}{1-\nu^2} \Delta
    x \, u_{z,i}, 
    \label{eq:single_spring}
\end{equation}
where $u_{z,i}$ is the local displacement of the $i$-th spring. The
penetration depth of the indenter is a function of its shape and of
the position of the tip of the end-effector. For example, for a flat
indenter, the spring deformation will be equal throughout the entire
contact area, whereas with a spherical indenter the spring in the
centre of the tip will be more compressed than the springs on the
sides.

\begin{figure}[t]
\centering
    \usetikzlibrary{matrix,fit} 
\usetikzlibrary{calc,patterns,decorations.pathmorphing,decorations.markings,positioning,backgrounds,arrows.meta,shapes,fit,matrix}
% \begin{document}

\begin{tikzpicture}[every node/.style={outer sep=0pt},thick,
 mass/.style={draw,thick},
 spring/.style={thick,decorate,decoration={zigzag,pre length=0.3cm,post
 length=0.3cm,segment length=6}},
 spring2/.style={thick,decorate,decoration={zigzag,pre length=0.3cm,post
 length=0.3cm,segment length=3, angle=10}},
 spring3/.style={thick,decorate,decoration={zigzag,pre length=0.3cm,post
 length=0.3cm,segment length=4, angle=10}},
 spring4/.style={thick,decorate,decoration={zigzag,pre length=0.3cm,post
 length=0.3cm,segment length=5, angle=10}},
 ground/.style={fill,pattern=north east lines,draw=none,minimum
 width=0.75cm,minimum height=0.3cm},
 dampic/.pic={\fill[white] (-0.1,-0.3) rectangle (0.3,0.3);
 \draw (-0.3,0.3) -| (0.3,-0.3) -- (-0.3,-0.3);
 \draw[line width=1mm] (-0.1,-0.3) -- (-0.1,0.3);}]

  \tikzstyle{damper}=[thick,decoration={markings,  
  mark connection node=dmp,
  mark=at position 0.5 with 
  {
    \node (dmp) [thick,inner sep=0pt,transform shape,rotate=-90,minimum width=15pt,minimum height=3pt,draw=none] {};
    \draw [thick] ($(dmp.north east)+(2pt,0)$) -- (dmp.south east) -- (dmp.south west) -- ($(dmp.north west)+(2pt,0)$);
    \draw [thick] ($(dmp.north)+(0,-5pt)$) -- ($(dmp.north)+(0,5pt)$);
  }}, decorate]

  % \node[mass,minimum width=6cm,minimum height=1cm,fill=red!70] (ft_sensor) {$ft_{sensor}$};
  \node[ground, minimum width=5.5cm, minimum height=3mm, anchor=north] (g1){};
  \node[above=1.5cm of g1, draw, circle, minimum size=3cm, anchor=south,fill=lightgray,thick] (cir){};
  \draw[line width = 0.5mm] (cir.-42) -- ++ (1.6,0) coordinate[right] (cr);
  \draw[thin,dashed] (cir.-42) -- ++ (-1.1,0) coordinate[midway] node[midway,above]{$a$};
  \draw[thin] (cir.center) -- (cir.134) coordinate[midway] node[midway,below]{$R_1$};
  \draw[line width = 0.5mm] (cir.222) -- ++ (-1.6,0) coordinate[midway](cl);
  \draw (g1.north west) -- (g1.north east);
  \draw[thin,dashed] (cr) -- ++ (0.75,0) coordinate[midway](ud);
  \draw[thin,dashed] (cr |- cir.south) -- ++ (0.75,0) coordinate[midway](ld);
  \draw[latex-] (ld) -- (ud) node[midway,right]{$d$};
  \draw[line width = 0.5mm] (cir.222) arc (222:318:1.5cm);
  \node[mass,above=5.5cm of g1,minimum width=5cm,minimum height=0.75cm,fill=red!70] (ft_sensor) {$\text{F/T}_\text{sensor}$};
  \draw[thick] (cir.0) -- (ft_sensor.east |- ft_sensor.south) coordinate[midway](l1); 
  \draw[thick] (cir.180) -- (ft_sensor.west |- ft_sensor.south) coordinate[midway](l2); 
        
    \draw[spring2] ([xshift=0mm]g1.north) coordinate(aux) 
    -- (aux|-cir.south) node[midway,right=1mm]{};
    \draw[spring3] ([xshift=5mm]g1.north) coordinate(aux)
   -- (aux|-cir.290) node[midway,right=1mm]{};
   \draw[spring4] ([xshift=10mm]g1.north) coordinate(aux)
   -- (aux|-cir.312) node[midway,right=1mm]{};
   \draw[spring] ([xshift=15mm]g1.north) coordinate(aux)
   -- (aux|-cr.south) node[midway,right=1mm]{};
   \draw[spring] ([xshift=20mm]g1.north) coordinate(aux)
   -- (aux|-cr.south) node[midway,right=1mm]{};
   \draw[spring] ([xshift=25mm]g1.north) coordinate(aux)
   -- (aux|-cr.south) node[midway,right=1mm]{};
    \draw[spring3] ([xshift=-5mm]g1.north) coordinate(aux)
   -- (aux|-cir.250) node[midway,right=1mm]{};
   \draw[spring4] ([xshift=-10mm]g1.north) coordinate(aux)
   -- (aux|-cir.228) node[midway,right=1mm]{};
   \draw[spring] ([xshift=-15mm]g1.north) coordinate(aux)
   -- (aux|-cl.south) node[midway,right=1mm]{};
   \draw[spring] ([xshift=-20mm]g1.north) coordinate(aux)
   -- (aux|-cl.south) node[midway](k1){};
   \draw[spring] ([xshift=-25mm]g1.north) coordinate(aux)
   -- (aux|-cl.south) coordinate[midway](k2)node[midway,left=1mm]{$\Delta k_z$};
    \begin{scope}[on background layer]
    {
        \filldraw[lightgray](ft_sensor.west|- ft_sensor.south)--(ft_sensor.east|- ft_sensor.south) -- (cir.0)--(cir.180);
    }
    \end{scope}
    \draw[thin,dashed](k1|-cl.north) --++ (0,0.8)coordinate[pos=0.85](aus1);
    \draw[thin,dashed](k2|-cl.north) --++ (0,0.8);
    \draw[latex-latex](k1|-aus1)--(k2|-aus1) node[midway, above=1mm]{$\Delta x$};
    \node at (0,5) {Indenter};
    
    \end{tikzpicture}

% \end{document}
    % \vspace{-0.8cm}
    \caption{Spherical indenter inside the elastic foundation. $d$ is the penetration, $a$ is the projection of the surface of the circle in contact, and $R_1$ is the equivalent radius of the 3D sphere.}
    \label{fig:contact}
\end{figure}
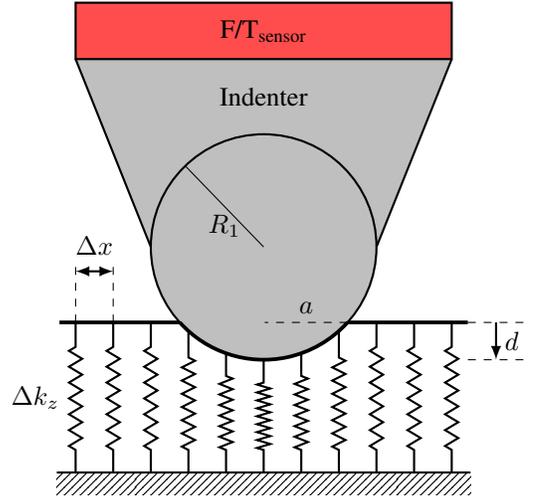

\noindent {\bf Spherical indenter.}
For a spherical indenter with radius $R$, the three dimensional
contact problem can be reformulated as an equivalent one-dimensional
problem, in which a linear elastic foundation is indented by a
circular indenter with radius $R_1 = R/2$ (see~\autoref{fig:contact} for reference).
%-%
%-%
Using the Hertzian theory, the circular profile of the indenter sphere
can be further approximated with a 2D parabolic interpolating profile
$g(x) = \frac{x^2}{2 R_1}$~\cite{johnson1985contact}.  More precisely,
let $d$ be the depth of penetration into the material. The force
exerted by an individual spring will vary according to its position
along the profile of the indenter. In the limit case of an
infinitesimal space between the springs, i.e., $\Delta x \simeq dx$,
the infinitesimal force contribution of a single spring element (as
given by equation \eqref{eq:single_spring}) can be rewritten as:
\begin{equation}
     df_{N} = \frac{E_f}{1-\nu^2} \left( d - \frac{x^2}{2 R_1} \right) dx.
     \label{eq:single_spring_inf}
\end{equation}
The portion of the indenter in contact with the surface is
$a = \sqrt{2 R_1 d}$, hence to obtain the total normal force generated
by the material, \eqref{eq:single_spring_inf} is integrated from $-a$
to $a$, obtaining
\begin{equation}
  \begin{aligned}
    F_N(d) &=  \int_{-a}^{a} df_N= \int_{-a}^{a} \left[ \frac{E_f}{1-\nu^2} \left( d -
          \frac{x^2}{2 R_1} \right) \right] dx = \\
    &=  \int_{-\sqrt{2 R_1 d}}^{\sqrt{2 R_1 d}}\left[
      \frac{E_f}{1-\nu^2} \left( d - \frac{x^2}{2 R_1} \right) \right]
      dx = \\
      &= \frac{4}{3} \frac{E_f}{1-\nu^2} d \sqrt{ 2 R_1 d} .
    \end{aligned}
    \label{eq:final_ext_contact}
\end{equation}
Substituting again $R_1 = R/2$, the result obtained
in~\eqref{eq:final_ext_contact} is the same as in the Hertzian
theory. % In Figure~\ref{fig:contact}, the one-dimensional model between
% a sphere and an elastic half-space is represented.

\noindent{\bf Generalisation to symmetric indenters. }
The previous formulation can be extended to any axially symmetric indenter. 
% For simplicity, we define the elastic half-space at $z=0$,
% thereby making $z$ the axis of symmetry. The surface of the indenter
% is then parameterised by the $x$ and $y$ coordinates.
% %
% For a generic axial symmetric indenter, the profile can be written as
% \begin{equation}
%   \Tilde{z} = l_n(r) = c_n r^n ,
%   \label{eq:3dprof}
% \end{equation}
% where $c_n$ is a constant depending on the shape of the profile,
% $r = \sqrt{x^2+y^2}$ and $n$ is an arbitrary positive number. 
Indeed, by defining the plane indenter geometry, the contact problem
is modelled as a 2D problem in combination with the linear elastic
foundation described above.  The reduced planar profile for a generic
axial symmetric shape can be expressed as
\begin{equation}
  g_n(x) = \tilde{c}_n |x|^n ,
\end{equation}
where $n$ is a generic positive number, and $\tilde{c}_n=k_n c_n$ is a constant, given by the product of a shape factor $c_n$ defining the shape of the 3D profile ($z = c_n r^n$, where $(r,z)$ are polar coordinates), and $k_n$ is a constant computed following the rule given by Hess~\cite{Popov2015MethodFriction}. Finally, the normal force can be
written as
\begin{equation}
  F_N(d) = \frac{2 n}{n +1 } \frac{E_f}{1 - \nu^2} \Tilde{c}_n^{-\frac{1}{n}} d^{\frac{n+1}{n}}.
    \label{eq:f_gen}
\end{equation}
%\dan{Are the forces $f_{N,i}$, $F_N$ and
%  $F_n$ the same? Please be consistent! }
% Using~\eqref{eq:f_gen}, I
It is possible to obtain the force equation
for the flat indenter and an indenter described by a function of degree four ($n=4$) as
\begin{align}
  & F_{N,\mathrm{flat}}(d) = \frac{E_f}{1 - \nu^2} 2 a d, \label{eq:flat} \\ 
  & F_{N,\mathrm{n=4}}(d) = \frac{8}{5} \frac{E_f}{1-\nu^2}  ( 2.667 c_4)^{-\frac{1}{n}} d^{\frac{5}{4}}.
  \label{eq:x4}
\end{align}

\subsection{Viscous Force Model}
\begin{figure}[t]
    \centering
    \includegraphics[]{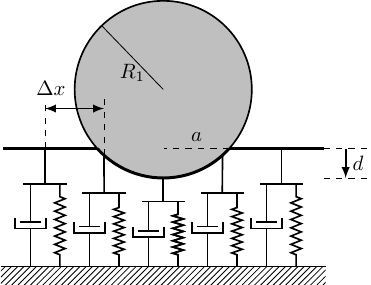}
    \caption{Schematic of the updated model containing the damper that describes the viscous behaviour of the material.}
    \label{fig:viscoelastic_model}
\end{figure}

The dimensionality reduction method can be extended to describe the
viscoelastic behaviour of the flexible substrate. Let us take the same
reduced system as in Fig. \ref{fig:contact} where the springs are
substituted by dampers. Popov et al.~\cite{Popov2015MethodFriction} have
shown that the expression of the force contribution of a single damper
has the following form
% similar to \eqref{eq:single_spring}, provided that 
% the penetration $u_{z,i}$ is replaced by  and the shear modulus is substituted with the viscosity modulus.
% Recalling that $\frac{E}{1-\nu^2}=\frac{2 G}{1 - \nu}$, the force generated by a single damper is equal to 
\begin{equation}
    \Delta f_{D,i} =  \frac{2}{1-\nu} \eta \Delta x\, \Dot u_{z,i},
    \label{eq:elastic_visc}
\end{equation}
where $\dot{u}_{z,i}$ is the penetration rate. Noting that, for a
generic indenter, ${u}_{z,i} = d- g_n(x)$, we can assume
$\dot{u}_{z,i}=\dot d$.  Hence, the force generated by a single
spring-damper (Kelvin-Voight) element (assumed infinitesimal, i.e.,
$\Delta x \simeq dx$) is the sum of~\eqref{eq:elastic_visc}
and~\eqref{eq:single_spring}, giving
\begin{equation}
    df_{\mathrm{DR},i} = \frac{E_f}{1-\nu^2} \left(d- g_n(x)\right) dx +  \frac{2}{1-\nu}  \eta \dot{d} dx.
    \label{eq:kv_el}
\end{equation}
\autoref{fig:viscoelastic_model} shows the schematic of a spherical
indenter in contact with a material modelled by multiple parallel
Kelvin-Voight elements.
%-%
%-%
Similarly to~\eqref{eq:final_ext_contact}, the force~\eqref{eq:kv_el}
can be integrated along the incontact surface of a spherical indenter
to get the total force as a function of the penetration into the
material, i.e.,
\begin{equation}
  \label{eq:TotalForce}
    F_{\mathrm{DR}}(d) =  \frac{4}{3} \frac{E_f}{1-\nu^2}  \sqrt{ R d}d +  \frac{4}{1-\nu}  \eta  \sqrt{ R d} \Dot d. 
\end{equation}
A similar result could be obtained considering an indenter with
generic shape, leading, of course, to a more complicated expression.

\subsection{Background on Offline Elasticity Estimation}
% \lui{Se questa parte e' background, come mi pare di cpaire, la foonderei con la seconda sezione intitolandola ``background on DR models and elasticity estimation'' così almeno è chiaro dove comincia la parte nuova}
As discussed in~\cite{BeberLPFSF24i2mtc}, if the contact force in
static conditions (i.e., at slow deformation rates that render viscous
forces negligible) is measured by a sensor mounted in series to the
indenter, a least square algorithm can be used to estimate the value
of the elasticity modulus that minimises
\begin{equation}
  \mathcal{L} = \sum_{j=i}^{L} \left(F_{\mathrm{sensor},j} - F_N(d_j) \right)^2,
  \label{eq:ls}
\end{equation}
where $L$ is the number of acquired measurement points, $F_{\mathrm{sensor},j}$ is the $j-$th force sample and $d_j$ is the $j-$th penetration value.
% , $n$ is the shape-factor discussed previously, and
% $\kappa$ is related to the elasticity as a function of the indenter
% shape. For example, using the sphere, $n=\frac{3}{2}$ and
% \begin{equation}
%     E_f = \kappa  \frac{3 ( 1 - \nu^2) }{ 4 \sqrt{R}} .
% \end{equation}
A challenge for the estimation problem is related to the difficulties
to locate the precise position of the specimen surface within the
robot frame, since the detection of the contact force is affected
by uncertainties. Therefore, 
% a constrained minimisation problem is
% defined on~\ref{eq:ls} as
the least square problem is cast by retaining the surface position as one of the unknowns and including a constraint on the value of the force:
\begin{equation}
    \begin{aligned}
    \mbox{argmin}_{E_f, z_{\mathrm{surf}}} \quad & \mathcal{L}\\
    \textrm{s.t.} \quad & -F_{\mathrm{unc}} \leq F(z_{\mathrm{surf}})\leq F_{\mathrm{unc}}.
    \end{aligned}
    \label{eq:minprob}
\end{equation}
where $z_{\mathrm{surf}}$ and $z_{\mathrm{EE}}$ are the $z$-axis positions of the
surface and of the end-effector, respectively, $d = z_{\mathrm{surf}} - z_{\mathrm{EE}}$
by definition, $F_{\mathrm{unc}}$ is the rated uncertainty of the sensor and
$F$ is the measured force.  This way, it is possible to find the
position of the surface with precision despite the uncertainties of
the force sensor, while at the same time estimating the elastic modulus $E_f$.

\section{Online Viscoelasticity Estimation Process}
This section presents an online algorithm that estimates penetration,
penetration rate, elasticity, and viscosity by modelling the contact
between a spherical indenter and a soft material using an Extended
Kalman Filter (EKF).

\subsection{Point Estimation} 
\label{subsec:PointEst}

% In~\cite{beber2024robotisedpalpationcancerdetection} is presented an EKF capable of estimating the viscoelastic parameters of a material by means of a static sinusoidal palpation.
To simplify the formulation of the force generated by a spherical
indenter presented in~\eqref{eq:TotalForce}, two new parameters have
been defined:
\begin{equation}
  \kappa  = \frac{4}{3} \frac{E_f}{1-\nu^2} \sqrt{R} \,\, \mbox{ and }\,\, \lambda  =  \frac{4}{1-\nu} \eta \sqrt{R},
\end{equation}
obtaining a new formulation of the total force~\eqref{eq:TotalForce}
as
\begin{equation}
        F_{DR}(d) =
        \begin{cases}
                \kappa d^{\frac{3}{2}} + \lambda   d^{\frac{1}{2}}
                \dot{d}, \;\; & d\geq0 ,\\
                0 \;\; & d<0.
        \end{cases}
        \label{eq:final_ext_contact2}
\end{equation}
A dynamical system can be obtained to describe the evolution of the
contact between the indenter and the soft surface given
in~\eqref{eq:final_ext_contact2}. First, we define the state vector
$\boldsymbol{x} = \left[d, \dot d, \kappa , \lambda \right]$,
describing the amount of penetration $d$, the penetration rate
$\dot d$, the normalised elasticity $\kappa$ and the normalised
viscosity $\lambda$. With $\Delta T$ being the sampling time of the
dynamic model and denoting with $x_{i,t}$ the $i$-th variable of
$\boldsymbol{x}$ at time instant $t \Delta T$, the discrete-time
system dynamics
$\boldsymbol{x}_{t+1}=f_t(\boldsymbol{x}_t,u_t,\boldsymbol{w})$
describing~\eqref{eq:final_ext_contact2} is then given by
\begin{equation}
  \begin{aligned}
    x_{1,t+1}&= x_{1,t} + \Delta T x_{2,t} +  \\&\frac{\Delta T^2}{2 m_I} \left(u_t -x_{1,t}^{\frac{3}{2}} x_{3,t} - x_{1,t}^{\frac{1}{2}}x_{2,t}x_{4,t}\right) + w_{1,t},\\
    x_{2,t+1}&= x_{2,t} + \frac{\Delta T}{m_I} \left(u_t -x_{1,t}^{\frac{3}{2}} x_{3,t} - x_{1,t}^{\frac{1}{2}}x_{2,t}x_{4,t}\right) + w_{2,t},\\
    x_{3,t+1}&= x_{3,t},\\
    x_{4,t+1}&= x_{4,t}.
  \end{aligned}
  \label{eq:state_space_NL_FT}
\end{equation}
The input $u_t$ is the force registered by a force-torque sensor along the
direction perpendicular to the surface, while
$\boldsymbol{w}_t = [w_{1,t}, w_{2,t}]^T$ is the i.i.d., zero mean,
Gaussian process model of the uncertainties, namely
$\boldsymbol{w}_t\sim\mathcal{N}(\mathbf{0},\mathbf{Q})$.
%with
%$\mathbf{Q}=\mathrm{diag}\left(
  % \sigma_{x_1}^2,\sigma_{x_2}^2\right)$. 
The input $u$ gives the uncertainty of the measurement model,
the force measured by the F/T sensor. The process noise
$\boldsymbol{w}_t$ can be written as a function of the standard
deviation $\sigma_{u}$ of the input force $u$. $\sigma_{u}$ can be
considered a time-invariant quantity, as it remains constant
throughout the experiment. The standard deviations of the penetration
and the velocity can be computed as follows:
\begin{equation}
     \sigma_{x_1} = \frac{\Delta T^2}{2 m_I} \sigma_{u} \,\, \mbox{ and }\,\,\sigma_{x_2} = \frac{\Delta T}{m_I} \sigma_{u}.
    \label{eq:model_unc}
\end{equation}
Said $\boldsymbol{B}=\left[\frac{\Delta T^2}{2 m_I} \ \ 
  \frac{\Delta T}{m_I}\right]^\intercal$, the covariance matrix
$\boldsymbol{Q}$ of the uncertainties $\boldsymbol{w}_t$ can then be
defined as
\begin{equation}
    \boldsymbol{Q} = \boldsymbol{B} \sigma_u^2 \boldsymbol{B}^T.
\end{equation}

The measurement function $z_t = h(\boldsymbol{x}_t,v_t)$ is instead
the velocity of the end effector in the direction orthogonal to the
surface, i.e.
\begin{equation}
        z_t = x_{2,t} + v_t,
        \label{eq:measurament}
\end{equation}
where $v_t$ is the i.i.d zero mean Gaussian process generating the
velocity measurement uncertainty,
$v_t\sim\mathcal{N}(0,\sigma_{vel}^2)$, which is the only uncertainty
contribution here.

Using the knowledge of the model of the system
$f_t(\boldsymbol{x}_t,u_t,\boldsymbol{w}_t)$, its input measurements
$u_t$ and the stochastic description of the associated uncertainties
$\boldsymbol{w}_t$, together with the velocity measurements
$h(\boldsymbol{x}_t,v_t)$, an EKF is employed to derive the contact
properties (penetration extent and velocity, elasticity and viscosity)
estimates. Similar to what was done
in~\cite{haddadi2012real,Schindeler2018OnlineLinearization}, the
palpation will be simulated by a sinusoidal motion perpendicular to
the soft surface.  In the interest of space, the EKF equations are not
reported in this manuscript, but a similar implementation can be found
in our previous
report~\cite{beber2024robotisedpalpationcancerdetection}.
% For more details about this implementation, please refer to~\cite{beber2024robotisedpalpationcancerdetection}.

\subsection{Dynamic Estimation} 
\label{subsec:DynamicEst}

Although point estimation is a useful method for measuring specific
properties, it is not a practical approach for mapping the stiffness
variation of a material, since an extensive number of measurement
points would be needed. Moreover, during a point estimation,
elasticity and stiffness remain constant throughout the
measurement phase, which is no longer the case for dynamic
estimation. In other words, the value at which the EKF based
on~\eqref{eq:state_space_NL_FT} converge for those parameters, which
are supposed to be constant due to a lack of model information,
actually change dynamically. Therefore, to increase the expressivity
of the model presented in Section~\ref{subsec:PointEst}, the system
described in~\eqref{eq:state_space_NL_FT} is expanded with a dynamic
term for the elasticity and viscosity values. More in depth, we
redefine the state as $\boldsymbol{x}\in\mathbb{R}^8$, with
$\boldsymbol{x}=\left[d, \dot d, \kappa , \lambda, \Dot \kappa, \Dot
  \lambda, \Ddot \kappa, \Ddot \lambda \right]$, whose dimension is twice as
much as the one in~\eqref{eq:state_space_NL_FT}.
Incorporating second-order derivatives as state variables is an effective strategy for accelerating
convergence, particularly in scenarios characterized by rapid variations in viscoelastic parameters.
This capability proves especially significant in real-time safety-critical applications, such as physical
human-robot interaction, where accurate estimation of viscoelastic parameters enables precise prediction of interaction forces exerted on a patient’s body, tailored to the specific anatomical location
involved.
The new discrete-time model will then be
\begin{equation}
    \begin{aligned}
      x_{1,t+1}&= x_{1,t} + \Delta T x_{2,t} +  \\&\frac{\Delta T^2}{2 m_I} \left(u_t -x_{1,t}^{\frac{3}{2}} x_{3,t} - x_{1,t}^{\frac{1}{2}}x_{2,t}x_{4,t}\right) + w_1,\\
    x_{2,t+1}&= x_{2,t} + \frac{\Delta T}{m_I} \left(u_t -x_{1,t}^{\frac{3}{2}} x_{3,t} - x_{1,t}^{\frac{1}{2}}x_{2,t}x_{4,t}\right) + w_{2},\\x_{3,t+1}&= x_{3,t} + \Delta T x_{5,t} + \frac{\Delta T^2}{2} x_{7,t},\\
      x_{4,t+1}&= x_{4,t} + \Delta T x_{6,t} + \frac{\Delta T^2}{2} x_{8,t},\\
      x_{5,t+1}&= x_{5,t} + \Delta T x_{7,t},\\
      x_{6,t+1}&= x_{6,t} + \Delta T x_{8,t},\\
      x_{7,t+1}&= x_{7,t},\\
      x_{8,t+1}&= x_{8,t},
    \end{aligned}
    \label{eq:state_space_NL_FT_plus}
\end{equation}
which is adopted as the model for the prediction step of the EKF.

However, the dynamics considered for the elasticity and viscosity
(i.e., the last six variables in~\eqref{eq:state_space_NL_FT_plus})
model a second-order system, which cannot account for abrupt changes,
as instead happens in presence of a mass in the body. Therefore, to
make the EKF more prompt and, hence, detect such rapid changes, we use
as a detection metric the {\em likelihood} between the estimated model
and the measurements. This entails defining an inflating fudge factor
$\alpha \in \left[1,\alpha_{\mathrm{max}}\right]$, which multiplies
the filter covariance matrix~\cite{bar2004estimation}. In instances
where the error exceeds a specified threshold $\theta$, $\alpha$
increases; conversely, it decreases. The increment and decrement are defined as $\Delta \alpha$, in the experimental section we will see the effect of varying this value. Letting
$\bar y_t = z_t - h(\Bar{\boldsymbol{x}}_{t|t-1})$ be the innovation,
where $\Bar{\boldsymbol{x}}_{t|t-1}$ is the predict state, the
updating rule of $\alpha$ is:
\begin{enumerate}
\item Calculate the magnitude of the innovation $\bar y_t$
  % \dan{This is not the innovation covariance, but just the innovation:
  %   what do you mean here?};
\item Update the fading factor \( \alpha \) based on the magnitude of
  the innovation: 
  % \dan{$\Delta \alpha$ is not defined}
  \begin{equation*}
    \alpha =
    \begin{cases}
      \min(\alpha_{max}, \alpha + \Delta \alpha), \;\; & \text{if } \bar y_t > \theta ;\\
      \max(1, \alpha - \Delta \alpha), \;\; & \text{otherwise}.
    \end{cases}
    % \label{eq:fading_factor_update}
  \end{equation*}
   %  \begin{align*}
   % &\text{If } \bar y_t > \theta:\\
   % &\alpha = \min(\alpha_{max}, \alpha + \Delta \alpha)\\
   % &\text{Else:}\\
   %  &\alpha = \max(1, \alpha - \Delta \alpha) 
   % \end{align*}
\item Correct the predicted covariance of the estimates as
  \begin{equation*}
    \boldsymbol{P}_{t|t-1} = \alpha \boldsymbol{P}_{t|t-1}.
  \end{equation*}
\end{enumerate}
Increasing the fading factor $\alpha_{max}$ will decrease the low pass
effect of the filter, decreasing the delay of the estimation process
but increasing the uncertainties. On the other hand, increasing the
threshold $\theta$ will have the opposite effect.
% \dan{Where is the updating rule
%   of $\theta$? Please report the updating rule here. Moreover, What is
%   the value of $\theta$ in the experiments? How it is chosen? What is
%   the rationale? Please report this discussion in the experimental
%   section.}
% This allows to increase the value of the covariance
% matrix when the value of the estimation and the measured value are too
% far apart.

The procedure for dynamic palpation is divided into two distinct
steps. In the initial phase, sinusoidal palpation is conducted in a
direction perpendicular to the surface of the material, with no
movement along the other axes. This is done to allow the filter
sufficient time to converge to the appropriate values of the
states. In the subsequent phase, sinusoidal palpation continues, but
movement in a direction of interest is also introduced. Notice that
different speeds of the end-effector may have different effects on the
elasticity estimates, therefore this aspect should be carefully
investigated.

% \dan{Ok: we have the model, but all the algorithm is missing. Where is
%   the estimator? Where is the analysis of the uncertainties? Where are
%   the details about the procedure and how the algorithm should be
%   applied? Are there specified trajectories? How defined? What is the
%   best trajectory to follow to estimate the parameters of interest?
%   Are there some simulations to show and to play with? If so, please
%   add a Section with the simulations.}

\section{Experimental Setup}
\label{sec:setup}
\begin{figure}[t]
  \centering
  \includegraphics[angle=-90,width=4.5cm]{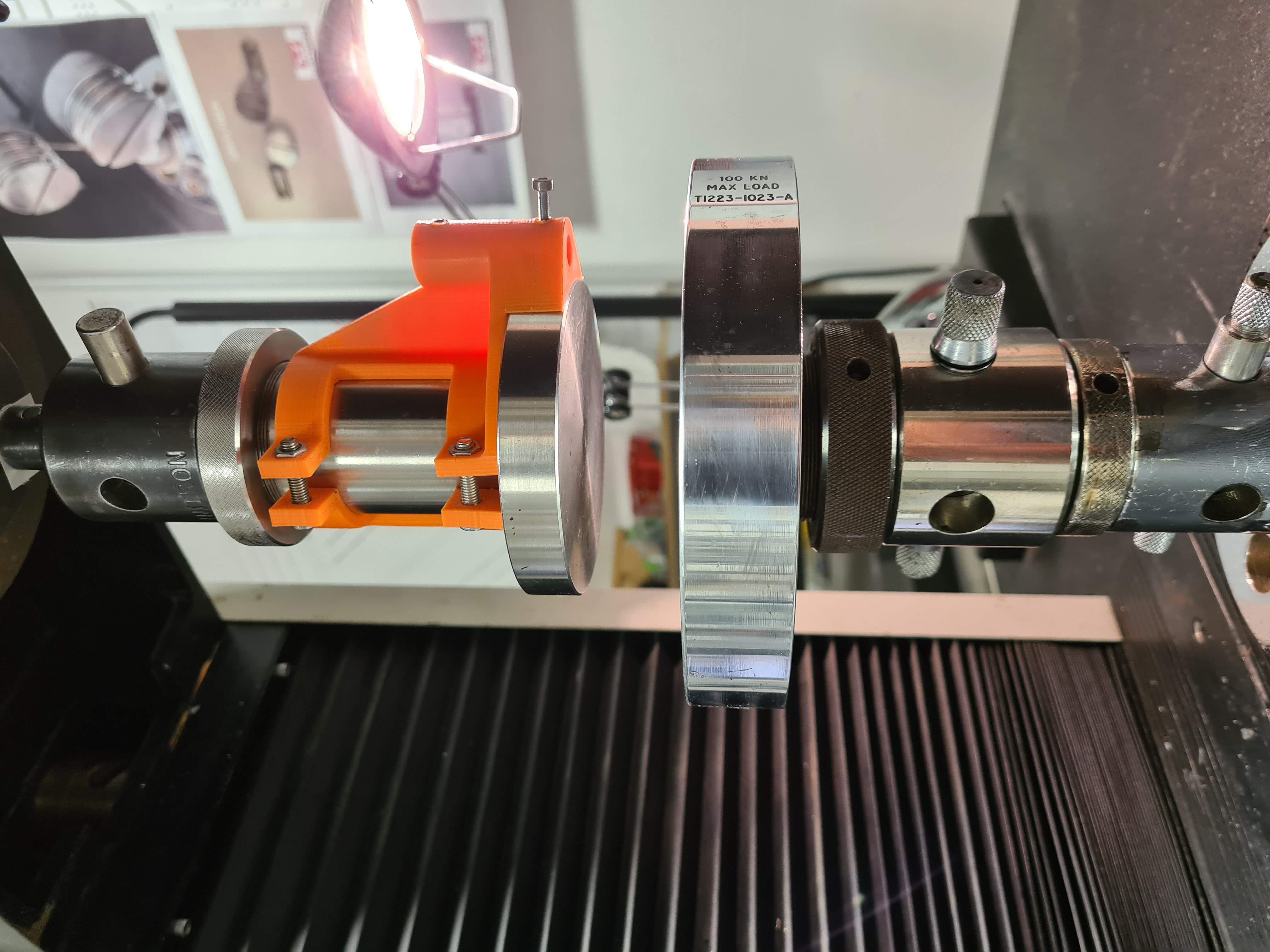}
  \caption{Instron® 4502 dynamometer used for the ground truth
    collection of the specimens characteristics.}
  \label{fig:dyn}
\end{figure}

The measurement system was characterized
using a set of specimens whose properties were determined with a
reference instrument for compression testing, serving as the ground truth. Specifically, quasi-static compression tests were
conducted at $23^\circ$C using an Instron® 4502 dynamometer (Norwood,
MA, USA) equipped with a $\SI{50}{\kilo \newton}$ load cell, as shown in
Fig.~\autoref{fig:dyn}. 
%-%
%-%
During the tests, the cross-head speed was of
$50$~mm/min and the sampling rate of the dynamometer was of $25$~pt/s
(or $25$~Hz) up to $5$~mm of compression.

\subsection{Robotic system}

The experimental setup includes an Ur3e, a widely used
collaborative $6$ degree-of-freedom robotic arm. It has a maximum
payload of $3$~kg and a pose repeatability per ISO 9283 of
$\pm 0.03$~mm. It operates at \SI{500}{\hertz}. At the end-effector of
the manipulator, it is attached a $6$-axis force-torque sensor, i.e.,
the BOTA System SensONE. The sensor works at \SI{1}{\kilo \hertz} with
standard deviation of the signal over 1 second of measurements in
stable conditions of \SI{0.05}{N} in the $z$-direction. After a
warm-up phase, the measurements can be considered uncorrelated since
there is a drift of less than \SI{1.5}{N/h} and our experiments have a
maximum duration of 40 seconds. Attached to the sensor there is a 3D
printed indenter. The robotic arm is controlled by a simple motion
controller in the Cartesian space. Hence, given the Cartesian position
of the end-effector as input, the values of the joints are calculated
by the robot inverse kinematics. A PD controller regulates the convergence speed, with the proportional gain empirically set to $20$ and the derivative gain to $0.5$, ensuring that the end-effector follows the trajectory with negligible error (i.e., below one millimeter)..

\subsection{Data Acquisition Protocol}

End-effector position and velocity data are obtained from the robot encoders using forward kinematics based on the Denavit-Hartenberg convention~\cite{Uigker1964AnMechanisms}. With these
rules, it is possible to obtain a homogeneous transformation matrix
for each of the $m$ joints as
\begin{equation}
  \label{eq:RigidTransf}
  \boldsymbol{T}_i = 
  \begin{bmatrix}
    \boldsymbol{R}_i & \boldsymbol{p}_i \\
    \boldsymbol{0} & 1
  \end{bmatrix} ,
\end{equation} 
where $i$ is the joint number,
$\boldsymbol{R}_i \in \mathbb{R}^{3\text{x}3}$ is the orientation
matrix representing the orientation of the joint $i$ and
$\boldsymbol{p}_i \in \mathbb{R}^{3\text{x}1}$ is the position of the
joint $i$. The pose of the end effector for the $m=6$ joint arm can
then be expressed as the product of the $m$ transformations
in~\eqref{eq:RigidTransf}, i.e.,
\begin{equation}
  \label{eq:RigidEE}
  \boldsymbol{T}_{ee} = \boldsymbol{T}_1 \boldsymbol{T}_2 \cdots
  \boldsymbol{T}_m .
\end{equation}
Similarly, it is possible to use the Jacobian matrix to obtain, from
the robot joint velocities, and the end effector velocities. While
position values are used within the offline
estimation~\eqref{eq:minprob}, speed values are used within the filter
measurement function~\eqref{eq:measurament}. In the proposed EKF, the
position, velocity, and acceleration data are expressed in millimeters
to avoid numerical errors, induced by the significant difference in
orders of magnitude between stiffness and damping, that may impair the
convergence of the filter.  The force data $u_t$ are collected
directly from the force-torque sensor mounted on the robot end-effector. They
are used both for the minimisation problem~\eqref{eq:TotalForce} and
as input for the filter prediction step
using~\eqref{eq:state_space_NL_FT_plus}. Using~\eqref{eq:model_unc}
and the value of the standard deviation of the force sensor along the
$z$ axis, it is possible to derive the model uncertainties
$\sigma_{x_1}^2 = \SI{1.12e-6}{m m}$ 
% \dan{add the missing information }
$\sigma_{x_2}^2 = \SI{0.3044}{m m/s}$ using a Type
B~\cite{therefore1995guide} derivation using~\eqref{eq:model_unc}.

Instead, the mass of the portion of the end effector that is placed
after the F/T sensor and in contact with the body has been estimated
by comparing the inertia of the system with the measured force during
a sinusoidal motion and using a least squares approach.  Finally, the
standard uncertainty on the velocity $\sigma_{vel}$ has been derived
by imposing on the robot end effector a sinusoidal motion along the
$z$-axis and then recording its cartesian velocity with a sampling
time of \SI{2}{m s}. Using a moving average filter of $21$ consecutive
samples, the root mean squared error between the filtered and
measured velocities is computed. This procedure has been done
$3$ times and, since the measurements had a negligible bias and the
sequence correlation in the time window of the experiments was again
negligible, we obtained a variance for the velocity measurements equal
to $\bar\sigma_{vel}^2 = \SI{0.4489}{m m/ s}$. 
% \dan{It is a velocity or a
%   distance? It is a variance or a standard deviation? It is better to
%   use the standard uncertainty (i.e., standard deviation) rather than
%   the veriance.}
%%% aggiungere come si stima l'incertezza sulla velocita'

% \dan{How this position is
%   useful to define the measurement function~\eqref{eq:measurament}?
%   What is the uncertainty on the computation of $d$? How this
%   uncertainties propagates to $v_t$ in~\eqref{eq:measurament}? How
%   evaluated?}. 
%   The force $u_t$ in~\eqref{eq:state_space_NL_FT_plus}
% generated by the material is instead equal to the force registered
% along the $z$-axis of the force torque sensor 
% \dan{What is the rated
%   uncertainty of the force sensor? How computed or retrieved? How this
%   uncertainty remaps on $w_1$ and $w_2$
%   in~\eqref{eq:state_space_NL_FT_plus}? Please report the uncertainty
%   analysis.}

\subsection{Specimens}
\label{subsec:specimens}

\begin{figure}
    \centering
    \includegraphics[width=0.95\linewidth]{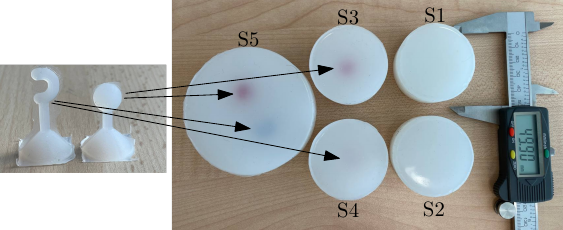}
    \caption{On the right are shown the cylindrical silicones used
      during the experiments, while on the left are the spherical and
      horseshoe-shaped intrusions.}
    \label{fig:silicones}
\end{figure}

Silicone specimens were used in this study due to their similarity to biological tissues~\cite{wells2011medical}, specifically ECOFLEX-0030, Dragonskin-10NV, and Dragonskin-30, listed from softest to hardest.. The specimens tested in the point-based experiments were cylinders with a
diameter of \SI{50}{\milli \meter} and a height of \SI{22}{\milli
  \meter}. Spherical and horseshoe-shaped inclusions were embedded in two samples made from the softer matrix to simulate stiffer cancerous tissue within softer, healthy tissue, as in~\cite{Yan2021}. Listed
below are the specimens used with their relative abbreviations:
\begin{itemize}
    \item \textbf{S1}: Dragonskin-10\SI{}{NV} matrix,
    \item \textbf{S2}: ECOFLEX-0030 matrix,
    \item \textbf{S3}: ECOFLEX-0030 matrix and Dragonskin-30 horseshoe intrusion,
    \item \textbf{S4}: ECOFLEX-0030 matrix and Dragonskin-30 spherical intrusion.
\end{itemize}
Finally, a bigger sample (\textbf{S5}), with diameter \SI{80}{m m},
containing both a spherical and horseshoe intrusion has been used for
the dynamic experiments. Specimens and intrusions are shown in~\autoref{fig:silicones}. We assumed the silicon samples, considering them
incompressible, with a $\nu = 0.5$. This is a broadly accepted assumption, which comes
from experimental evidence and is used in the vast majority of works
dealing with continuum mechanics of rubber-like
materials~\cite{wells2011medical, darby2022modulus}.

% \begin{figure}[t!]
%     \centering
%     \input{images/Tikz/siliconses}
%     \caption{Caption}
%     \label{fig:silicones}
% \end{figure}

\section{Elastic Modulus Estimation Process}
\label{sec:experiments}
% The tests for the elasticity module ground truth were performed on parallelepiped-shaped specimens (see
% Fig.~\ref{fig:cubetti}) with $50\times50$~mm section and thickness
% in range $20-30$~mm.
% %-%
% \begin{figure}[t]
%     \centering
%     \includegraphics[width=0.6\columnwidth]{images/photos/cubetti.jpg}
%     \caption{Specimens used during the experiments made of multilayer polymeric foams.}
%     \label{fig:cubetti}
% \end{figure}
% %-%
% The samples used in the experiments were polymeric foams made of
% polyethylene and formed from multiple layers to achieve the desired
% thickness for the experiments, which present elasticity values comparable with biological tissues~\cite{wells2011medical}. The geometric and physical
% characteristics are specified in Table~\ref{tab:cube_param}.
% %-%
% \begin{table}[t]
%     \centering
%     \caption{Geometric and physical parameters of the specimens.}
%     \label{tab:cube_param}
%     \begin{tabular}{ccc}
    
%     \textbf{Specimen} & \textbf{Density} [$\SI{}{\gram / \centi\meter^3}$] & \textbf{Length x Width x Height} [$\SI{}{\milli\meter^3}$]\\

%     \textbf{White}   & $0.0350$ & $50 \times 50 \times 30$ \\
    
%     \textbf{Pink}   & $0.0181$ & $50 \times 50 \times 20$ \\
    
%       \textbf{Grey}   & $0.0236$ & $50 \times 50 \times 20$ \\
    
%     \end{tabular}

% \end{table}

\begin{figure}[t!]
  \centering
  \includegraphics[]{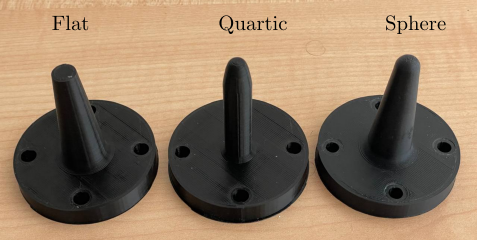}
  \caption{3D-printed tips utilised in the experiments. From left: the
    flat, the quartic and the spherical tips.}
  \label{fig:tips}
\end{figure}

% The tests carried out with the laboratory instrument in
% Fig.~\ref{fig:cubetti}, treated here as ground truth, have been
% repeated using the Ur3e manipulator. 
Experiments to estimate the elastic modulus with the UR3e
manipulator were conducted with three different tips: a flat tip with
\SI{5}{m m} radius, a spherical tip with \SI{5}{m m} radius and a
quartic surface with $c_4 = 10^7$ (see\eqref{eq:x4}), all shown in~\autoref{fig:tips}.
%-%
%-%
To compare the different tips, the penetration velocity remained
consistent at $50$~mm per minute. The indentation depth is adjusted to
$10\%$ of the sample thickness to stay within the range of material
linearity~\cite{Qiang2011EstimatingThickness}, since any deeper
penetration could amplify the non-linear characteristics of the
material that are not modelled. During the experiment, the position
and the force in the $z$-direction is collected for post-processing,
while~\ref{eq:minprob} is used to estimate the elasticity modulus.
%as described in Fig.~\ref{fig:schematic}. Using~\eqref{eq:minprob},
% We were able to reconstruct the force quite precisely with a mean
% residual value that never exceeded $\SI{0.03}{\kilo \pascal^2}$ for
% the spherical and paraboloidal tip and $\SI{0.06}{\kilo\pascal^2}$ for
% the flat tip, as can be seen in Table~\ref{tab:res_exp}.
% %-%
% \begin{table}[t]
%     \centering
%     \caption{Residual resulting from the force reconstruction $\left[\SI{}{\kilo \pascal^2}\right]$.}
%     \label{tab:res_exp}
%     \begin{tabular}{cccc}
    
%     \textbf{Specimen} & \textbf{Flat Tip} & \textbf{Sphere Tip} & \textbf{Paraboloid Tip} \\

%     \textbf{White} & 0.0316 & 0.0240 &  0.0292  \\
    
%     \textbf{Pink} & 0.0225 & 0.0250 & 0.0291 \\
    
%     \textbf{Grey} & 0.0613 & 0.0208 & 0.0308 \\
                    
%     \end{tabular}

\begin{figure}[t!]
    \centering
    \includegraphics{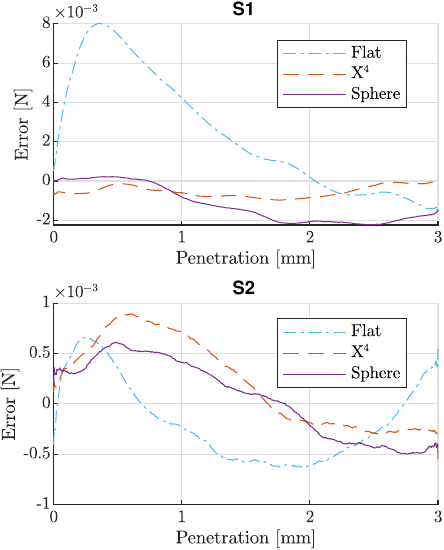}
    \caption{Average force error between the measured and the
      estimated value in the {\bf S1} and {\bf S2} specimens. 
    % In all experiments, the average error using the spherical and the paraboloid indenter is bounded, demonstrating the suitability of the model with these indenter shapes. On the other hand, the oscillating average error of the flat indenter does not ensure precise force estimation. %It can be seen that we were able to reconstruct the force accurately for the spherical and paraboloidal tips, but not for the flat tip.
    }
    \label{fig:forces}
\end{figure}

% \end{table}
% %-%
% The position of the surface for each specimen was estimated with
% relatively high accuracy except for the flat tip, resulting in a mean
% error for the spherical and paraboloidal tip of $0.38$~mm and for
% $0.39$~mm, respectively. Instead, since the flat tip had non-linearity
% of the phase after contact, we used the force measurements after
% $20\%$ of penetration to estimate the specimen elasticity.
\autoref{fig:forces} shows the mean error between the reconstructed
force using, respectively, the models
in~\eqref{eq:final_ext_contact},~\eqref{eq:flat} and~\eqref{eq:x4},
and the measurement force acquired by the F/T Sensor. More precisely
\begin{equation}
  \label{eq:ForceError}
   e_i =  F_{\mathrm{EST},i} - F_{\mathrm{FT},i},
\end{equation}
where $i$ is the i-th test, $F_{\mathrm{FT}}$ is the
measured force and $F_{\mathrm{EST}}$ is the estimated force with the different models. 
% The
% error was then further filtered with a low pass filter to remove the noise of the force sensor and keep only the behaviour of the error \dan{filtered how? To
%   what extent?}. 
The discrepancy is consistently minimal, except for
the flat indenter with the hardest silicone. The considerable initial
discrepancy, which is not discernible in the graph of the softer
silicone, can be attributed to the fact that the surfaces are not
perfectly parallel, resulting in a non-linear effect of force until
complete contact is achieved.

The estimated elasticity quantities are consistent with the ground
truth, as depicted in Table~\ref{tab:elast_table}.
%-%
\begin{table*}[t]
    \centering
    \caption{Results of the elastic modulus estimations of {\bf S1} and
      {\bf S2} using the three tips of different shape, with $50$
      tests each. $\mu$ is the elastic modulus, $\sigma$ is the standard
      uncertainty, and $\mathrm{Error}$ the relative error with
      respect to the ground truth.}
    % \vspace{-0.25cm}
    \label{tab:elast_table}
    \usetikzlibrary{matrix,fit} 

% \begin{document}
\definecolor{airforceblue}{rgb}{0.36,
0.54, 0.66}
\definecolor{cyan}{rgb}{0.3010, 0.7450, 0.9330}
\definecolor{orange}{rgb}{0.8500, 0.3250, 0.0980}
\definecolor{magenta}{rgb}{0.4940, 0.1840, 0.5560}
\begin{tikzpicture}[
  head color/.style args={#1/#2}{
    row 1 column #1/.append style={nodes={fill=#2}}},
  % swap order of row and column styles
  % matrix/inner style order={
  %   every cell,
  %   row, even odd row,
  %   column, even odd column,
  %   cell}
]

\matrix[
   matrix of nodes, nodes in empty cells,
   nodes={minimum width=1.5cm, align=center,
          minimum height=2em, anchor=center},
   % add striped row style
   row 2/.style={nodes={fill=airforceblue!50}},
   row 3/.style={nodes={fill=airforceblue!50}},
   row 4/.style={nodes={fill=airforceblue!50}},
   % modify the feature column and header row
   column 1/.style= {nodes={fill=airforceblue, inner ysep=0}},
   row 1/.style= {nodes={text depth=0.2ex, text=black}},
   row 1 column 1/.style={nodes={fill=none, draw=none}},
   head color/.list={2/teal!70,3/teal!70,4/cyan!70,5/cyan!70,6/cyan!70,7/orange!70,8/orange!70,9/orange!70,10/magenta!70,11/magenta!70,12/magenta!70} % specify header colors
  ] (m)
  {
     % &\shortstack{Ground Truth\\Elasticity $[\SI{}{\kilo\pascal}]$} & \shortstack{Ground Truth\\SD $[\SI{}{\kilo\pascal}]$}&\shortstack{Flat Tip\\Elasticity $[\SI{}{\kilo\pascal}]$} & \shortstack{Flat Tip\\SD $[\SI{}{\kilo\pascal}]$}&\shortstack{Spherical Tip\\Elasticity $[\SI{}{\kilo\pascal}]$} & \shortstack{Spherical Tip\\SD $[\SI{}{\kilo\pascal}]$}&\shortstack{Paraboidal Tip\\Elasticity $[\SI{}{\kilo\pascal}]$} & \shortstack{Paraboidal Tip\\SD $[\SI{}{\kilo\pascal}]$} \\

      & $\mu$ $[\SI{}{\kilo\pascal}]$ & $\sigma$ $[\SI{}{\kilo\pascal}]$ & $\mu$ $[\SI{}{\kilo\pascal}]$ & $\sigma$ $[\SI{}{\kilo\pascal}]$ & $\mathrm{Error}$ $[\%]$ & $\mu$ $[\SI{}{\kilo\pascal}]$ & $\sigma$ $[\SI{}{\kilo\pascal}]$ & $\mathrm{Error}$ $[\%]$& $\mu$ $[\SI{}{\kilo\pascal}]$ & $\sigma$ $[\SI{}{\kilo\pascal}]$ & $\mathrm{Error}$ $[\%]$ \\
      $\mathbf{S1}$ & $282.1$ & $0.3$   &   $302.3$   & $0.7$ & $7.9$ & $311.0$   &  $2.1$    & $11.07$  & $283.9$ & $0.9$ & $1.4$\\
      $\mathbf{S2}$ & $112.5$ & $0.2$   & $131.7$     & $0.6$ & $17.6$ & $122.1$   & $0.4$     & $8.9$  & $117.5$ & $0.5$&  $4.9$\\
      % \sc{Grey} & $194$ & $17$   & $227.46$     & $7.35$ & $-17.24$ & $202.16$   & $20.29$     & $-5.88$  & $173.97$ & $19.61$ & $10.32$ \\
      };

% &Ground Truth&Flat Tip & Spherical Tip & Paraboidal Tip\\
    \node[text depth=0.2ex, text=black, fill= teal!70, anchor=south west,minimum width=3cm, align=center, minimum height=2em] at (m-1-2.north west) {\sc{Ground Truth}};
    \node[text depth=0.2ex, text=black, fill= cyan!70, anchor=south west,minimum width=4.5cm, align=center, minimum height=2em] at (m-1-4.north west) {\sc{Flat Tip}};
    \node[text depth=0.2ex, text=black, fill= orange!70, anchor=south west,minimum width=4.5cm, align=center, minimum height=2em] at (m-1-7.north west) {\sc{Quartic Tip}};
    \node[text depth=0.2ex, text=black, fill= magenta!70, anchor=south west,minimum width=4.5cm, align=center, minimum height=2em] at (m-1-10.north west) {\sc{Spherical Tip}};
\end{tikzpicture}
% \end{document}
    % \vspace{-0.8cm}
\end{table*}
%-%
%-%
%-%
The flat tip exhibits the poorest behaviour with a maximum relative
error of nearly $20\%$, with a big difference between S1 and S2. This
could be, again, caused by the alignment imperfections between the
indenter. The estimates provided by the quartic indenter exhibits a
consistent error of approximately $10\%$ for both specimens. In
contrast, the spherical indenter exhibits the smallest error, with a
value below $5\%$.

\section{Viscoelasticity Experiments}
\label{sec:vc_experiments}
We start this section by experimentally comparing the chosen DR model
presented in Section~\ref{sec:models} with the two most widely adopted
alternatives in robotics, which are typically employed to represent
contact behaviours with soft bodies. The first is the Kelvin-Voigt
(KV) model, which is used for its simple linear representation that
ensures the contact forces description with a relatively minimal
degree of error. It is described by
\begin{equation}
  F_{KV}(d) =
  \begin{cases}
    K d + B
    \dot{d}, \;\; & d\geq0 ,\\
    0, \;\; & d<0 ,
  \end{cases}
  \label{eq:kv_eq}
\end{equation}
where $K$ is the stiffness and $B$ is the damping related to the
spring-damper system.  The second is the Hunt-Crossley (HC) model
described by
\begin{equation}
  F_{HC}(d) =
  \begin{cases}
    K_c d^{n} + B_c d^{n}
    \dot{d}, \;\; & d\geq0 ,\\
    0, \;\; & d<0 ,
  \end{cases}
  \label{eq:hc_eq}
\end{equation}
where $K_c$ is the stiffness, $B_c$ is the damping and $n$ is a
parameter dependent on the type of
contact~\cite{Pappalardo2016Hunt-CrossleySurgery}. This non-linear
model is capable of faithfully reconstructing the contact force and
does and widely reduces the representation error
of~\eqref{eq:kv_eq}. The HC model also has the advantage of not being
defined for any specific shape of the contact surface. This
flexibility comes with a main drawback: the coefficient $n$
in~\eqref{eq:hc_eq} must be estimated online, which makes the
estimation process more challenging. Another disadvantage is the
absence of a direct relationship between the stiffness and the
elasticity modulus of the material, and between the damping value and
the viscosity modulus.

\begin{figure}[t]
    \centering
    \resizebox{0.9\columnwidth}{!}{\begin{tikzpicture}
    [spy using outlines={lens={scale=2,rotate=-16}, ellipse, size=2.5cm, height=1.5cm, connect spies},
    ]
        \node[inner sep=0pt] (robot) at (0,0) {\includegraphics{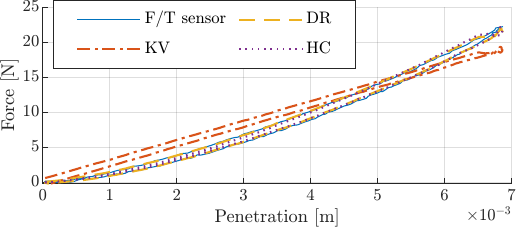}};
    
        \spy [red] on (0.4,0) in node [left,lens={scale=2,rotate=2},rotate=16] at (4,0.2);

    \end{tikzpicture}}
  \caption{Reconstruction of the force of the load tests using the $3$
    KV, HC and DR models. $n = 1.53$ is estimated for the
    HC. }
    \label{fig:loadcycle}
    %\vspace{-3mm}
\end{figure}

\begin{figure}[t]
    \centering
    \includegraphics[scale=1]{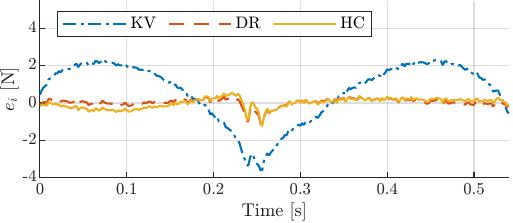}
    \caption{Error between the force measurements and the force
      computed by the contact models.}
    \label{fig:error_comp}
\end{figure}

A load test was conducted to evaluate the performance of the three
models (KV, HC and DR) under varying indentation speeds. This test
involved a load and unload phase at a constant speed. In this phase,
the elasticity and viscosity parameters, including the exponent $n$ of the HC model
in~\eqref{eq:hc_eq}, are computed offline using a least squares
minimisation. \autoref{fig:loadcycle} shows the result of these
minimisations.
% -%
% -%
The evolution of the error $e_i$ computed as in~\eqref{eq:ForceError},
is shown in~\autoref{fig:error_comp}.
% -%
% -%
As anticipated, the KV model is suitable whenever an accurate force
reconstruction is not needed. This model does not take into account the variations of the contact surface during indentation. Its load cycle is also energetically inconsistent~\cite{diolaiti2005contact}, as the force does not start and end
at zero since the contact velocity is not zero. Both the HC and the DR method turn
to be effective for the force representation.  The HC model presents an additional layer of complexity due to the nonlinearity of the unknown
exponent that has to be estimated online during the estimation. For a quantitative
comparison, the residual values of the least squares are reported in
Table~\ref{tab:model_residuals}.
%-%
\begin{table}[t]
\renewcommand{\arraystretch}{1.2}
\caption{Residuals resulting from the least-squares minimisation
  process employed to estimate the parameters derived from the load
  tests.}
%\vspace{-2mm}
\centering
    \begin{tabular}{c c c}
    \toprule
\multicolumn{3}{c}{\sc{Residual Value [\SI{}{N^2}]}}\\
        \midrule
        Kelvin-Voight & Dimensionality-Reduction & Hunt-Crossley\\
        & & $\beta$ = 1.53\\ 
        \cline{3-3} 
        792.45 & 13.62 & 20.59\\
        \bottomrule
    \end{tabular}
\label{tab:model_residuals}
%\vspace{-4mm}
\end{table}
It can be observed that, despite the residual errors of the HC and DR
models are similar, the latter exhibits smaller errors (see~\autoref{fig:error_comp}).

The residuals with different indentation speeds have been also
computed and reported in Table~\ref{tab:model_residuals2}, adopting
the elasticity and viscosity values obtained from the previous test.
%-%
\begin{table}[t]
\renewcommand{\arraystretch}{1.2}
\caption{Residuals with varying indentation speed.}
%\vspace{-2mm}
\centering
    \begin{tabular}{c c c}
    \toprule
\multicolumn{3}{c}{\sc{Residual Value [\SI{}{N^2}]}}\\
        \midrule
        Velocity & Dimensionality-Reduction & Hunt-Crossley\\
        & & $\beta$ = 1.53\\ 
        \cline{3-3} 
        $v_0$ & 13.62 & 20.59\\
        $v_1$ & 20.18 & 32.08\\
        $v_2$ & 20.14 & 23.75\\
        \bottomrule
    \end{tabular}
\label{tab:model_residuals2}
%\vspace{-4mm}
\end{table}
% -%
This additional check was performed to ensure that no overfitting in
the first load cycle data is present. It is possible to observe that
the results obtained with the initial tests at speed
$v_0=\SI{6}{m m /s}$ match the tests at speeds $v_1=\SI{4}{m m/s}$ and
$v_2 = \SI{8}{m m /s}$, proving that the adopted models are
correct. The experimental evidence shows that the adopted DR model is
the best fit for the problem at hand.

\subsection{Online Point Estimation}
\label{subsec:OnlinePointEst}

We report here on the online point estimation carried out with an EKF
based on the DR model of~\eqref{eq:state_space_NL_FT}
and~\eqref{eq:measurament}, and used to estimate the elasticity and
viscosity online. At the end-effector is imposed a sinusoidal motion
along the $z$-axis, which is perpendicular to the sample and equal to
\begin{equation}
  z(t) = z_0 + z_a \sin(2\pi \omega t),
\end{equation}
where $z_0$ is the bias inside the sample, i.e., the initial
penetration, $z_a$ is the amplitude of the sinusoid, and $\omega=\SI{2}{Hz}$ is
the frequency of the palpation. The selection of palpation frequency constitutes a trade-off between the limitation of robot dynamic
properties, the precision of elasticity estimation, and viscosity estimation. Frequencies that are
excessively high result in the estimation of erroneous elasticity values, while frequencies that are
insufficiently low preclude the estimation of the viscous properties of the material. The EKF starts when the indenter is in
contact with the surface. $z_0$ was set to \SI{2}{m m} for the
smallest tip and to \SI{4}{m m} for the other two tips, while $z_a$
was set to \SI{1}{m m} regardless of the tip shape. These choices
ensure motion in the linear elastic regime of the samples, since a
bigger penetration would cause the bulging of the lateral surface. The
filter based on~\eqref{eq:state_space_NL_FT} is initialised with an
initial state $\boldsymbol{x}_0 = \left[1,v_0,0.1,0.01\right]$, where
\SI{1}{m m} is the initial guess of the penetration, $v_0$ is the
initial velocity along the $z$-axis measured by the robot, $0.1$ is
the initial guess on $\kappa$ and $0.01$ on $\lambda$. The covariance
matrix $\boldsymbol{P}$ of the EKF is initialised as a diagonal matrix
$\boldsymbol{P}=\mbox{diag}(\left[5,1,1,1\right])$, whose values have been
tuned through experimental tests.

\subsubsection{Estimates Correlation}
\label{sec:corr}
\begin{figure}[t]
    \centering
    \includegraphics[width=0.9\linewidth]{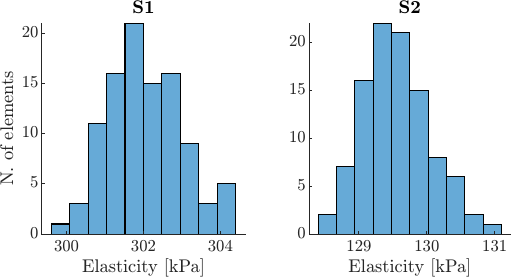}
    \caption{Histogram of the elasticity measurements of the hard
      {\bf S1} and soft {\bf S2} silicone samples.}
    \label{fig:hist_sil}
\end{figure}

\begin{figure}[t]
    \centering
    \includegraphics[width=0.95\linewidth]{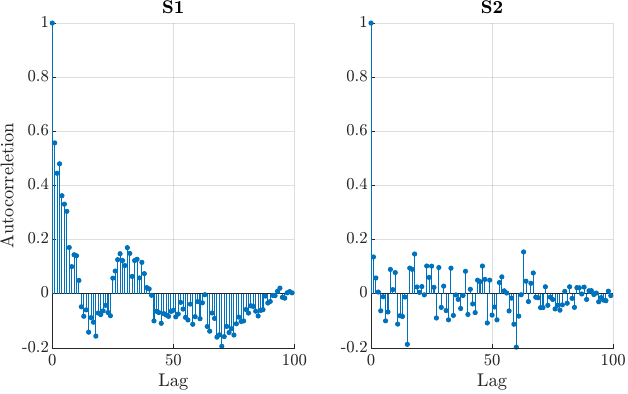}
    \caption{Autocorrelation function of the elasticity measurements
      of the hard {\bf S1} and soft {\bf S2} silicone samples.}
    \label{fig:autocorr}
\end{figure}

\begin{figure}[t]
  \centering
  \includegraphics[width=0.9\linewidth]{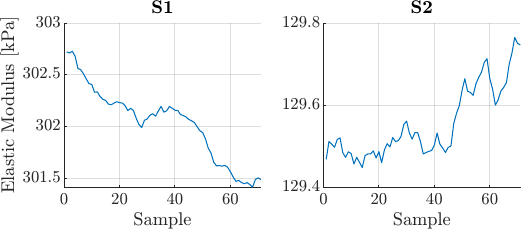}
  \caption{Moving average of the elasticity measurements for the hard
    {\bf S1} and soft {\bf S2} silicone samples with a sliding window
    of $30$ elements.}
  \label{fig:moving_av}
\end{figure}

To ensure that the elasticity of the sample does not change along the
experiment execution due to the relaxation effect (i.e., the effect
that changes the viscoelastic characteristics of a material after an
excitation), $n = 100$ measurements were taken for the hard {\bf S1}
and soft {\bf S2} silicone samples and reported in~\autoref{fig:hist_sil} with a histogram.
%-%
We then computed the correlation coefficients $\rho_{\bf S1} = 0.19$
and $\rho_{\bf S2} = -0.45$ for the hard and soft silicone,
respectively,
% \begin{align*}
%     &r_{\mathrm{soft}} = -0.07  &r_{\mathrm{hard}} = 0.5
% \end{align*}
which substantiates the existence of a slight correlation between the
data of the hard silicone and soft silicon. Those values are clearly shown in the
autocorrelation function plot of~\autoref{fig:autocorr}.
%-%
%-%
To further investigate this trend, we calculated the moving average
using a sliding window. We utilise a window size of $30$ values, which
smoothes out short-term fluctuations and highlights longer-term
trends. From~\autoref{fig:moving_av}, it can be seen that there is
a slight decrease for S1 and increase for S2 in the stiffness when the number of measurements
increases. For our experiments, this effect is
deemed to be negligible.
%-%
%-%

\subsubsection{Elastic modulus estimate Validation}
The aim here is to perform a comparison between the online estimate of
the elastic modulus and the value obtained through a compression test
performed with the dynamometer Instron® 4502 described in
Section~\ref{sec:setup} and depicted in~\autoref{fig:dyn}, whose
measurements are considered as ground truth. The detailed procedure
to obtain such quantities is described in
Appendix~\ref{app:correction_factor}. To investigate the impact of
varying tip diameter on estimation accuracy, three spherical tips with
diameters $r$ of \SI{4}{m m}, \SI{10}{m m} and \SI{20}{m m} were
adopted. For each tip, $50$ measurements were obtained and filtered
with the EKF~\eqref{eq:state_space_NL_FT} and then compared with $10$ measurements taken with the compression test. Table~\ref{tab:elast_table2}
collects the results of the tests thus described: the $r=2$~mm tips yielded accurate elasticity estimates for both the soft
and hard samples. Increasing the size of the tip also increases the error in the estimate. This phenomenon can be attributed to the DR model's assumption of infinite flat surface contact, which fails to account for the potential effects that may arise when this assumption is invalid. Notably, the effects are limited and consistent for both silicones as long as the tip radius is not excessively large, i.e., 5 mm. However, when employing a 10 mm tip, these effects become too invasive, and the estimate becomes invalid. 

However, one can consider using a 5mm radius tip despite the absolute error in elasticity estimation because in a scouting setup it covers a larger area and is therefore more likely to provide useful information than a very small tip such as the 2mm radius tip.
%-%
\begin{table*}[t]
    \label{tab:elasticty_silicones}
    \centering
    \caption{Results of the elastic modulus estimations of {\bf
        S1} and {\bf S2} using the three spherical tips along $50$
      tests each. $\mu$ is the mean value of the elasticity, $\sigma$
      is the standard uncertainty, and $\mathrm{Error}$ the relative
      error with respect to the ground truth (i.e., the $10$ averaged
      dynamometer measurements).}
    % \vspace{-0.25cm}
    \label{tab:elast_table2}
    \usetikzlibrary{matrix,fit} 

% \begin{document}
\definecolor{airforceblue}{rgb}{0.36,
0.54, 0.66}
\definecolor{cyan}{rgb}{0.3010, 0.7450, 0.9330}
\definecolor{orange}{rgb}{0.8500, 0.3250, 0.0980}
\definecolor{magenta}{rgb}{0.4940, 0.1840, 0.5560}
\begin{tikzpicture}[
  head color/.style args={#1/#2}{
    row 1 column #1/.append style={nodes={fill=#2}}},
  % swap order of row and column styles
  % matrix/inner style order={
  %   every cell,
  %   row, even odd row,
  %   column, even odd column,
  %   cell}
]

\matrix[
   matrix of nodes, nodes in empty cells,
   nodes={minimum width=1.5cm, align=center,
          minimum height=2em, anchor=center},
   % add striped row style
   row 2/.style={nodes={fill=airforceblue!50}},
   row 3/.style={nodes={fill=airforceblue!50}},
   % modify the feature column and header row
   column 1/.style= {nodes={fill=airforceblue, inner ysep=0}},
   row 1/.style= {nodes={text depth=0.2ex, text=black}},
   row 1 column 1/.style={nodes={fill=none, draw=none}},
   head color/.list={2/teal!70,3/teal!70,4/cyan!70,5/cyan!70,6/cyan!70,7/orange!70,8/orange!70,9/orange!70,10/magenta!70,11/magenta!70,12/magenta!70} % specify header colors
  ] (m)
  {
     % &\shortstack{Ground Truth\\Elasticity $[\SI{}{\kilo\pascal}]$} & \shortstack{Ground Truth\\SD $[\SI{}{\kilo\pascal}]$}&\shortstack{Flat Tip\\Elasticity $[\SI{}{\kilo\pascal}]$} & \shortstack{Flat Tip\\SD $[\SI{}{\kilo\pascal}]$}&\shortstack{Spherical Tip\\Elasticity $[\SI{}{\kilo\pascal}]$} & \shortstack{Spherical Tip\\SD $[\SI{}{\kilo\pascal}]$}&\shortstack{Paraboidal Tip\\Elasticity $[\SI{}{\kilo\pascal}]$} & \shortstack{Paraboidal Tip\\SD $[\SI{}{\kilo\pascal}]$} \\

      & $\mu$ $[\SI{}{\kilo\pascal}]$ & $\sigma$ $[\SI{}{\kilo\pascal}]$ & $\mu$ $[\SI{}{\kilo\pascal}]$ & $\sigma$ $[\SI{}{\kilo\pascal}]$ & $\mathrm{Error}$ $[\%]$ & $\mu$ $[\SI{}{\kilo\pascal}]$ & $\sigma$ $[\SI{}{\kilo\pascal}]$ & $\mathrm{Error}$ $[\%]$& $\mu$ $[\SI{}{\kilo\pascal}]$ & $\sigma$ $[\SI{}{\kilo\pascal}]$ & $\mathrm{Error}$ $[\%]$ \\
      $\mathbf{S1}$ & $282$ & $0.3$ & $276.8$   &  $1.4$    & $1.8$  &   $302.4$   & $0.8$ & $7.2$   & $377.1$ & $1.3$ & $33$\\
      $\mathbf{S2}$ & $112$ & $0.2$  & $109.7$   & $1.1$     & $2.2$  & $129.5$     & $0.5$ & $15.7$  & $175.8$ & $0.8$&  $57$\\
      };

% &Ground Truth&Flat Tip & Spherical Tip & Paraboidal Tip\\
    \node[text depth=0.2ex, text=black, fill= teal!70, anchor=south west,minimum width=3cm, align=center, minimum height=2em] at (m-1-2.north west) {\sc{Ground Truth}};
    \node[text depth=0.2ex, text=black, fill= cyan!70, anchor=south west,minimum width=4.5cm, align=center, minimum height=2em] at (m-1-4.north west) {\sc{Spherical Tip} $r=\,$\SI{2}{m m}};
    \node[text depth=0.2ex, text=black, fill= orange!70, anchor=south west,minimum width=4.5cm, align=center, minimum height=2em] at (m-1-7.north west) {\sc{Spherical Tip} $r=\,$\SI{5}{m m}};
    \node[text depth=0.2ex, text=black, fill= magenta!70, anchor=south west,minimum width=4.5cm, align=center, minimum height=2em] at (m-1-10.north west) {\sc{Spherical Tip} $r=\,$\SI{10}{m m}};
\end{tikzpicture}
    % \vspace{-0.8cm}
\end{table*}
%-%

\subsubsection{Viscosity Validation}

As the compression tests were conducted with the dynamometer in
Section~\ref{sec:setup} which performs quasi-static tests, it is not
possible to extrapolate the viscous values of the silicone
specimens. Consequently, we relatively compare the viscosity values
obtained through the online estimation process among them with
different spherical tip radii. The viscosity estimates, collected in
Table~\ref{tab:visc_silic}, reveals that the \SI{2.5}{m m} and
\SI{5}{m m} radius tips are highly similar, whereas those for the
\SI{10}{m m} tips are quite different, thus confirming the test outputs
collected in Table~\ref{tab:elast_table2}.
%-%
\begin{table}[t]
\label{tab:viscosity_silicones}
\renewcommand{\arraystretch}{1.2}
\caption{Comparisons of the average $\mu$ and the standard deviation
  $\sigma$ of the viscosity estimates for {\bf S1} and {\bf S2}
  samples with a spherical tip of different radii.}
\centering
    \begin{tabular}{c | c c | c c | c c }
      \toprule
      \multicolumn{7}{c}{\sc{Viscosity Estimates [\SI{}{P s}]}} \\
      \midrule
      & \multicolumn{2}{c}{$r=\,$\SI{2}{m m}} & \multicolumn{2}{c}{$r=\,$\SI{5}{m m}} & \multicolumn{2}{c}{$r=\,$\SI{10}{m m}}  \\
      \cline{2-7} 
      Sample & $\mu$ & $\sigma$ & $\mu$ & $\sigma$ & $\mu$ & $\sigma$ \\ 
      \hline
      {\bf S1} & 1028 & 36 & 860 & 35 & 1000 & 33 \\
      {\bf S2} & 517 & 25 & 362 & 14 & 444 & 16\\
      \bottomrule
    \end{tabular}
\label{tab:visc_silic}
\end{table}
%-%
Hence, from this point on the rest of the experiments will be
conducted with the $r=5$~mm spherical indenter.

\subsubsection{Lump Detection}
The point online estimation was then used to identify the viscoelastic
parameters of specimens containing harder silicone parts, as in cancer detection palpation tests. Hence, the objective of
this experimental study was to see if there is a change in stiffness
and if this change is significant enough to be used in a detection
process. Table~\ref{tab:el_silic2} reports the results of these
experiments: both harder insertions can be detected by looking at the
mean $\mu$ of the elasticity values. The statistical analysis, conducted using the Student's t-test, confirmed that the differences in tumour detection results were highly significant, with p-values falling well below the 0.01 threshold.
%-%
\begin{table}[t]
\renewcommand{\arraystretch}{1.2}
\caption{Comparisons of the average $\mu$ and the standard deviation
  $\sigma$ of the elasticity and viscosity estimates for the modified
  specimens {\bf S3} and {\bf S4}, described in
  Section~\ref{subsec:specimens} and compared to the value without
  hard insertions.}
%\vspace{-2mm}
\centering
    \begin{tabular}{c | c | c c | c c }
      \toprule
      \multicolumn{6}{c}{\sc{Elasticity and viscosity estimates}} \\
      \midrule
      & \multicolumn{1}{c}{{\bf S2}} & \multicolumn{2}{c}{{\bf S3}} &
                                                                      \multicolumn{2}{c}{{\bf S4}}  \\
      \cline{2-6} 
      Modulus & $\mu$ & $\mu$ & $\sigma$ & $\mu$ & $\sigma$ \\ 
      \hline
      Elasticity [\SI{}{k Pa}] & 129.5 & 142.8 & 0.7 & 171.4 & 1.0 \\
      Viscosity [\SI{}{Pa  s}] & 362 & 399 & 16 & 458 & 16 \\
      \bottomrule
    \end{tabular}
\label{tab:el_silic2}
%\vspace{-4mm}
\end{table}
%-%
Moreover, the low standard deviation exhibited in these experiments
opens to an efficient detection process.

\subsection{Online Dynamic Estimation}

\begin{figure}[t]
    \centering
    \includegraphics[trim={0cm 6.3cm 1.2cm 1.5cm},clip, width=0.9\linewidth]{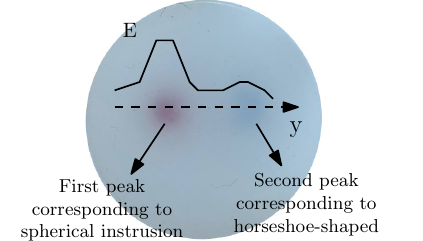}
    \caption{Trajectory (y) of the robot end effector on the silicon sample containing the spherical and horse-shaped intrusions.}
    \label{fig:traj}
\end{figure}

\begin{figure}[t]
    \centering
    \includegraphics[width=0.9\linewidth]{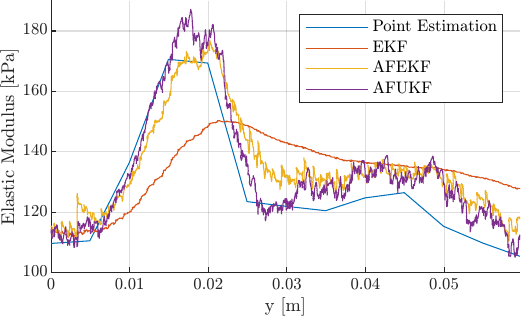}
    \caption{Dynamic elasticity estimation using the three different
      filters adopted, i.e., EKF, AFEKF, and AFUKF, with a comparison
      with the point estimation, which is made by sampling the silicone
      once every $5$~mm (that is, the indenter radius). The two picks
      in the figure refer to the sphere and the horseshoe beneath the
      silicone surface.}
    \label{fig:model-comp}
\end{figure}

The transition from point to dynamic palpation increases the efficacy
and speed of the analysis but presents a more challenging estimation
scheme since the indenter tip now has to slide on the surface while
ensuring a continuous palpation with desired contact forces. In order
to test the effectiveness of this new dynamic scheme, the results are
compared with the point estimation approach of~\autoref{subsec:OnlinePointEst} when the estimation points are
chosen sufficiently close to each other and, hence, resulting in a
sampling based ground truth for the dynamic estimators at hand. The
additional variables of the dynamic estimators in $\boldsymbol{x}_0$,
which is adopted in~\eqref{eq:state_space_NL_FT_plus}, are initialised
to zero since at the beginning the elasticity and viscosity parameters
will be constant with a covariance of
$\mbox{diag}(\boldsymbol{P})=\left[5,1,1,1,0.1,0.1,0.1,0.1\right]$.
The {\bf S5} specimen was employed in these experiments: as discussed
in~\autoref{subsec:specimens}, this specimen has two embedded
intrusions, one is a sphere and the other one a horseshoe (see~\autoref{fig:silicones} and~\autoref{fig:traj}). The continuous test starts with a
$10$~second point palpation, as in the previous test, and then
progresses in a predetermined direction at a constant speed. The
outcomes of the continuous palpation are illustrated in~\autoref{fig:model-comp}.
%-%

%-%
Here, the continuous estimation is performed employing three different
filtering techniques: a) the EKF presented previously, to see how the estimation works without any improvement; b) the Adaptive
Fading Extended Kalman Filter (AFEKF)~\cite{KimLPJ07icca}, which adopt
a forgetting factor on the covariance matrix computation to
account for modelling uncertainties in this parameter; c) the Adaptive
Fading Unscented Kalman Filter (AFUKF)~\cite{HuGZS15ijacsp} that uses the expanded state and a fading scheme similar to the AFEKF but applied to the Unscented
Kalman Filter (UKF). The fading factors $\theta$ and $\alpha$ adopted
in filters have been discussed in detail in~\autoref{subsec:DynamicEst}. Therefore, the three compared
solutions are:
\begin{itemize}

\item The EKF using equations~\eqref{eq:state_space_NL_FT}
  and~\eqref{eq:measurament};

\item The AFEKF using again equations~\eqref{eq:state_space_NL_FT}
  and~\eqref{eq:measurament};

\item The AFUKF using the expanded state
  in~\eqref{eq:state_space_NL_FT_plus} and the measurement function
  in~\eqref{eq:measurament}.
  
\end{itemize}

A comparison has been made between the elasticity estimation of the
three models and the punctual palpation performed with $5$~mm
intervals, which is equal to the radius of the indenter. It is
expected that two elasticity picks are detected: the first when the
harder silicone sphere is encountered beneath the surface; the second
for the horseshoe. From Fig.~\ref{fig:model-comp}, the EKF cannot
estimate the correct values due to the slow convergence rate, while
the AFEKF accurately estimates the elasticity values but still
exhibits a slight delay. The AKUKF eliminates the delay in estimation
but introduces greater oscillation and a larger overshoot when the
indenter encounters a stiffer region. Both the filters that use the
adapting fading factor can identify two picks, with the second being
more pronounced in the UKF case.

\begin{figure}[t]
    \centering
    \includegraphics[width=0.9\linewidth]{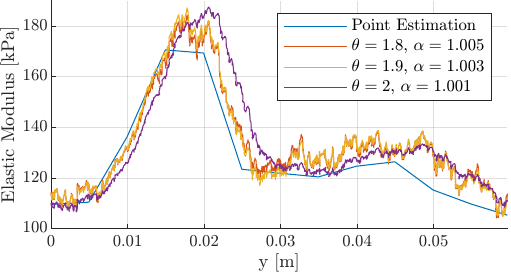}
    \caption{Effect of varying fading values $\theta$ and $\alpha$ on
      the elasticity estimates compared, with a comparison with the
      point estimation.}
    \label{fig:fading_effect}
\end{figure}

\begin{figure}
    \centering
    \includegraphics[width=0.9\linewidth]{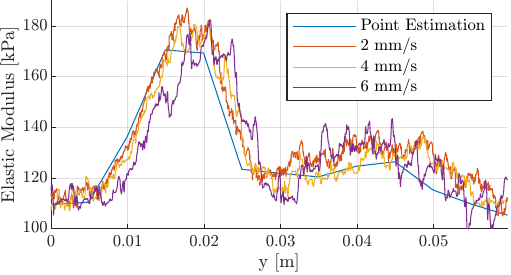}
    \caption{Effect of indenter velocity (in mm/s) on the elasticity
      estimates, again compared with the point
      estimation.}
    \label{fig:velocity_diff}
\end{figure}

To further investigate the impact of the fading factor on the best
performing AKUKF dynamic estimation algorithm,
Fig.~\ref{fig:fading_effect} presents a performance comparison when
the fading factors $\theta$ and $\alpha$ change.
%-%
%-%
With an increased value of $\theta$ and lower of $\alpha$, the
estimates are smoother but the responsiveness is reduced. Conversely,
reducing $\theta$, the filter is more responsive but the estimation
exhibits significant uncertainty. Consequently, a
compromise between these two behaviours can be determined by direct
calibration. Moreover, the AFUKF was tested with varying indenter
velocities to determine how the estimation performance would
change. The results are reported in Fig.~\ref{fig:velocity_diff}.
%-%
%-%
It was found that both the $2$~mm/s and $4$~mm/s velocities showed
good performance and a high degree of similarity, while the $6$~mm/s
velocity had an increased estimation uncertainty.

\begin{figure}[t!]
    \centering
    \begin{adjustbox}{width=.5\linewidth,center}
      \input{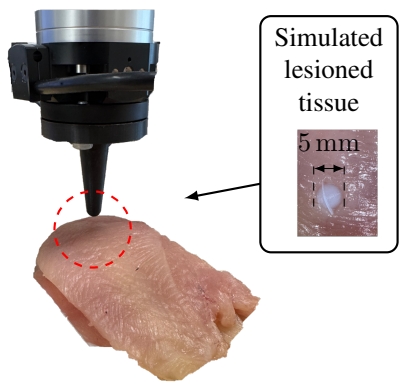}
    \end{adjustbox}
    \caption{Proof-of-concept experiment with chicken breast. On the
      right, it is possible to see the silicon sphere used to simulate
      the damaged tissue during the viscoelasticity estimation.}
    \label{fig:chick_exp_rr}
\end{figure}

\subsection{Preliminary Test on Ex Vivo Biological Tissue}
    
To provide a preliminary evaluation of the method on biological
tissue, we conducted a proof-of-concept ex vivo experiment with
chicken breast~\cite{othman2022stiffness,
Qiang2011EstimatingThickness}. The aim of this test is to extend
the results with silicones to biological sample, assessing the
system ability to estimate the viscoelastic properties of soft
biological material, both in the presence and absence of a stiffer
inclusion, which was created by inserting a \SI{5}{mm}-diameter
silicone sphere made of DragonSkin-30, identical to the one used in
the S4 scenario, between two chicken breast slices, each
approximately \SI{15}{mm} thick. To accommodate the sphere, a slot
was cut into the upper surface of the lower slice using a scalpel,
as shown in~\autoref{fig:chick_exp_rr}. In the first trial, the
robot palpated the tissue without any inclusion to establish
baseline viscoelastic properties. In the second trial, the silicone
sphere was inserted between the slices to simulate the presence of a
stiffer lesion within soft tissue.

\begin{table}[t!]
  \caption{Estimated viscoelastic properties (mean and standard
    deviation) of ex vivo chicken breast samples with and without a
    stiffer silicone inclusion over 10 repeated palpations.}
\label{tab:chicken_viscoelasticity}

\centering
    \begin{tabular}{c | c c | c c }
      \toprule
      \multicolumn{5}{c}{\sc{Elasticity and viscosity estimates}} \\
      \midrule
       & \multicolumn{2}{c}{{\bf Chicken Brest}} &
                                                                      \multicolumn{2}{c}{{\bf Chicken Brest with Sphere}}  \\
      \cline{2-5} 
      Modulus & $\mu$ & $\sigma$ & $\mu$ & $\sigma$ \\ 
      \hline
      Elasticity [\SI{}{k Pa}] & 53.7 & 5.2 & 102.7 & 10.7 \\
      Viscosity [\SI{}{Pa\;s}] & 623 & 101 & 1095 & 150 \\
      \bottomrule
    \end{tabular}
\end{table}

The results, summarised in Table~\ref{tab:chicken_viscoelasticity},
show a clear difference between the two conditions. The presence of
the stiffer inclusion resulted in a noticeable increase in both
estimated elasticity and viscosity. These findings indicate that the
proposed method can effectively detect and localise mechanical
anomalies, such as stiffer inclusions, within biological tissue. The
higher standard deviations observed in this experiment, compared to
those obtained with homogeneous silicone samples~\ref{tab:elasticty_silicones},~\ref{tab:viscosity_silicones},~\ref{tab:el_silic2}, highlight the
natural variability of biological materials and the challenges in
maintaining consistent contact conditions across repeated
measurements.

Despite these non-ideal measurement conditions, the results demonstrate promising potential for clinical
applications, particularly in identifying and localising stiffer
regions within soft tissue, such as tumours or fibrotic areas. The
method’s ability to estimate viscoelastic properties through robotic
palpation offers valuable diagnostic support, especially in contexts
where visual feedback is limited, such as minimally invasive or
robot-assisted procedures. Although preliminary, the successful
application to ex vivo tissue confirms the feasibility of using the
method on biological material and lays the groundwork for future
integration into clinical studies.

\section{Conclusion}

An method to estimate the viscoelastic parameters of a soft
material using a robotic arm has been presented. This is relevant
to the development of robotic medical applications based on the
physical interaction between a probe and the patient's body, but can
also be used for many other applications where the interaction is of
fundamental importance. In the first set of experiments, after offline force reconstruction, we validated our model by comparing
it with FT sensor data and benchmarked it against existing viscoelastic models. We showed that it is possible to
reconstruct the contact force with high precision using different
types of indenters and with more precision than it is currently
possible using the Hunt-Crossley model. In the second set of experiments, we used an EKF to estimate online the point-wise elasticity of the
silicone samples and compared the results with those from a compression test where the elasticity was
computed offline. A comparison of the elasticity
data obtained by compression testing and those obtained by indentation
revealed an error of no more than $3\%$ with respect to the adopted
reference and with the spherical tip. This error is sufficiently small to differentiate between healthy and diseased tissue, which usually differ by at least one order of magnitude. To illustrate this point, consider the example of healthy breast tissue, its elastic modulus is 20 times less than the modulus of diseased tissue~\cite{paszek2005tensional,xu2023role}. Indeed, two types of rigid lumps, rendering cancerous inclusion in a tissue, were successfully identified, with one causing a $14\%$
variation in elasticity. In the third set of experiments, continuous viscoelastic estimation was performed and compared with
point-wise estimations, confirming the method’s accuracy in dynamic palpation scenarios. We have shown that through dynamic
estimation, although not highly precise in determining the absolute elastic modulus values,
we were able to detect its very small variations. It has to be noted
that the problem of the incorrect absolute elasticity estimates can be solved
by checking the points of interest with the point estimation approach.

Future work will concentrate on integrating the dynamic estimation technique into a search algorithm
for the location of tumours. Furthermore, it is planned to extend the tests from silicones to both healthy
and diseased biological tissues, thus enabling comprehensive validation of this framework in the
targeted use case of tumour detection and characterization.

% or


\appendices
\section{Ground truth elastic modulus measurements}
\label{app:correction_factor}
% For silicon samples, the elasticity value with a compression test can not be calculated solely on the slope of the stress-strain curve. 
% Given that these materials are incompressible when compressed, they undergo a displacement of the outer surface, which alters their material response. To address this issue, a finite element model has been developed to incorporate this effect and extract a correction factor for obtaining the actual elasticity value from the calculated value based on the slope,

Ground truth values for the elastic modulus used for comparison in
Table \ref{tab:elast_table2} were obtained performing compression
tests on the cylindrical samples {\bf S1} and {\bf S2} with an Instron
4502 dynamometer. Given that the silicone materials under
investigation are incompressible ($\nu =0.5$)~\cite{wells2011medical, darby2022modulus}, the elastic modulus
$E_f$ cannot be simply obtained as the slope of the measured
stress-strain curves, as the samples undergo complex deformations
characterised by a bulging of their lateral surface. To extrapolate
the value of the elastic modulus from the measurements, we performed a
finite element simulation using Comsol Multiphysics (Nonlinear
Structural Materials module). We used a one-parameter Neo-Hooke
incompressible hyperelastic model and simulated the compression
behaviour of a sample with the same aspect ratio (diameter-to-height
ratio) as the tested specimens. In the simulations, we assumed that
the transverse bases of the samples are fixed in the radial direction
and can only move relative to one another in the axial direction,
whereas the lateral surface is free. Using the results from the FEM
simulations, we found that for samples with diameter-to-height ratio
of $2.32$ the following stress-strain relationship holds:
\begin{equation}\label{eq:correction_factor}
    P \simeq 1.75 E_f \varepsilon
\end{equation}
where $P$ is the nominal stress on the sample (force per unit nominal
area) and $\varepsilon$ is the lumped axial strain (displacement over
initial height).  The ground truth value of the elastic modulus $E_f$
was obtained by fitting the data to the trend
in~\eqref{eq:correction_factor}.

% \section{}

\section*{Acknowledgment}
This research has been funded by the MUR ``Departments of Excellence
2023-27'' program (L.232/2016), by the PNRR project FAIR - Future AI
Research (PE00000013), and by the European Union projects INVERSE (GA
no. 101136067) and MAGICIAN (GA no. 101120731).
%This research has been funded by the MUR ``Departments of Excellence 2023-27'' program (L.232/2016), by the PNRR project FAIR - Future AI Research (PE00000013), and by the European Union projects INVERSE (grant agreement no. 101136067) and MAGICIAN (grant agreement no. 101120731).

\ifCLASSOPTIONcaptionsoff
  \newpage
\fi



%\bibliographystyle{IEEEtran}


% \end{thebibliography}
% \balance
\bibliographystyle{IEEEtran}
\bibliography{references_TIM}
% \vspace{-10mm}
%\vskip -1\baselineskip plus -1fil

\begin{IEEEbiography}[{\includegraphics[width=1in,height=1.25in,clip,keepaspectratio]{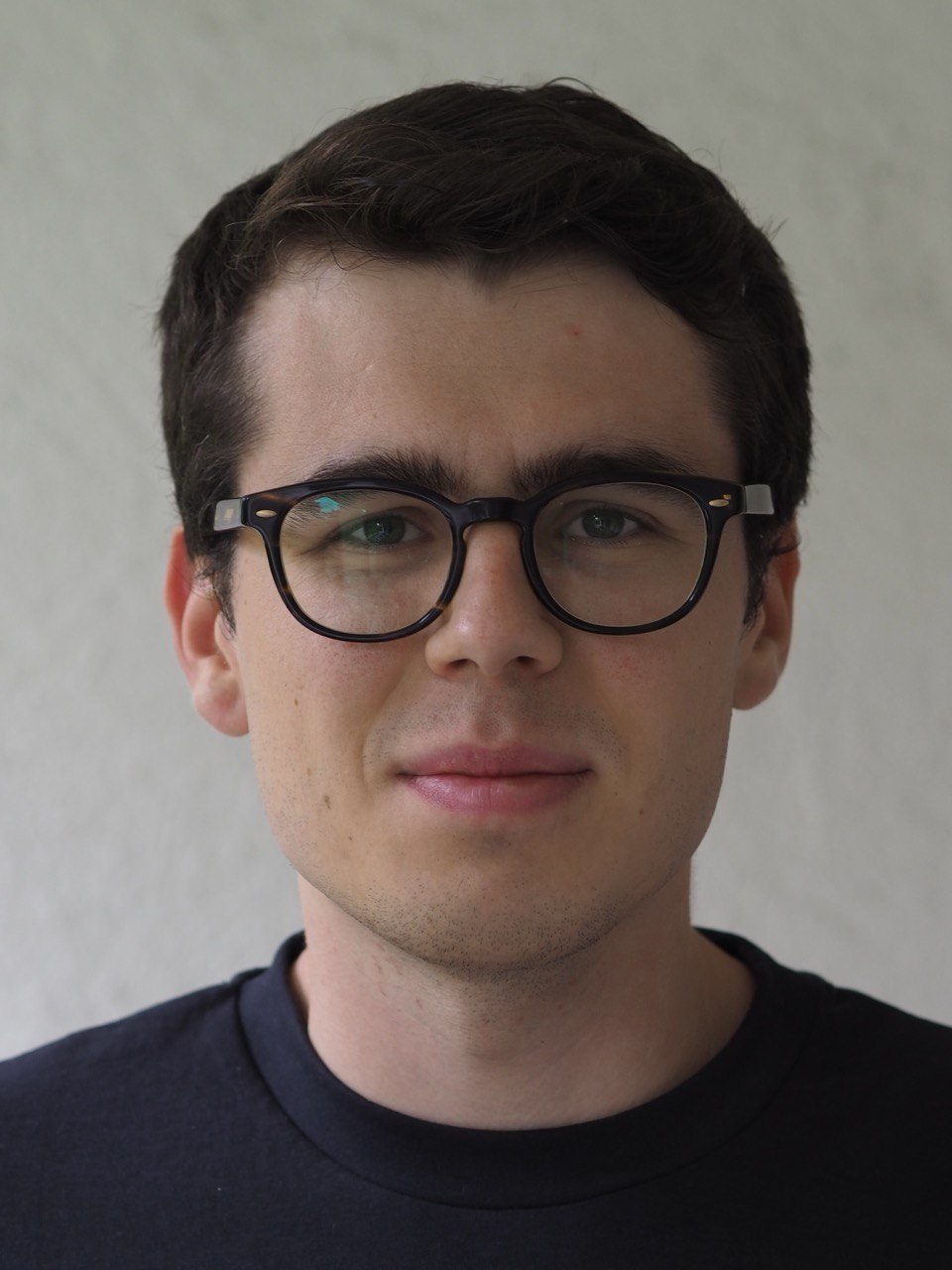}}]{Luca Beber} 
 received his M.Sc. in Mechatronics Engineering in 2023. He is currently pursuing a Ph.D. within the DRIM doctoral programme in the Assistive and Medical Robotics of the University of Trento, focusing on control and motion planning of medical robotic arms for ultrasound and palpation examinations. His research interests include robotic manipulation, force-based interaction, and the development of robotic systems for medical applications.
\end{IEEEbiography}

\vskip -2\baselineskip plus -1fil

\begin{IEEEbiography}[{\includegraphics[width=1in,height=1.25in,clip,keepaspectratio]{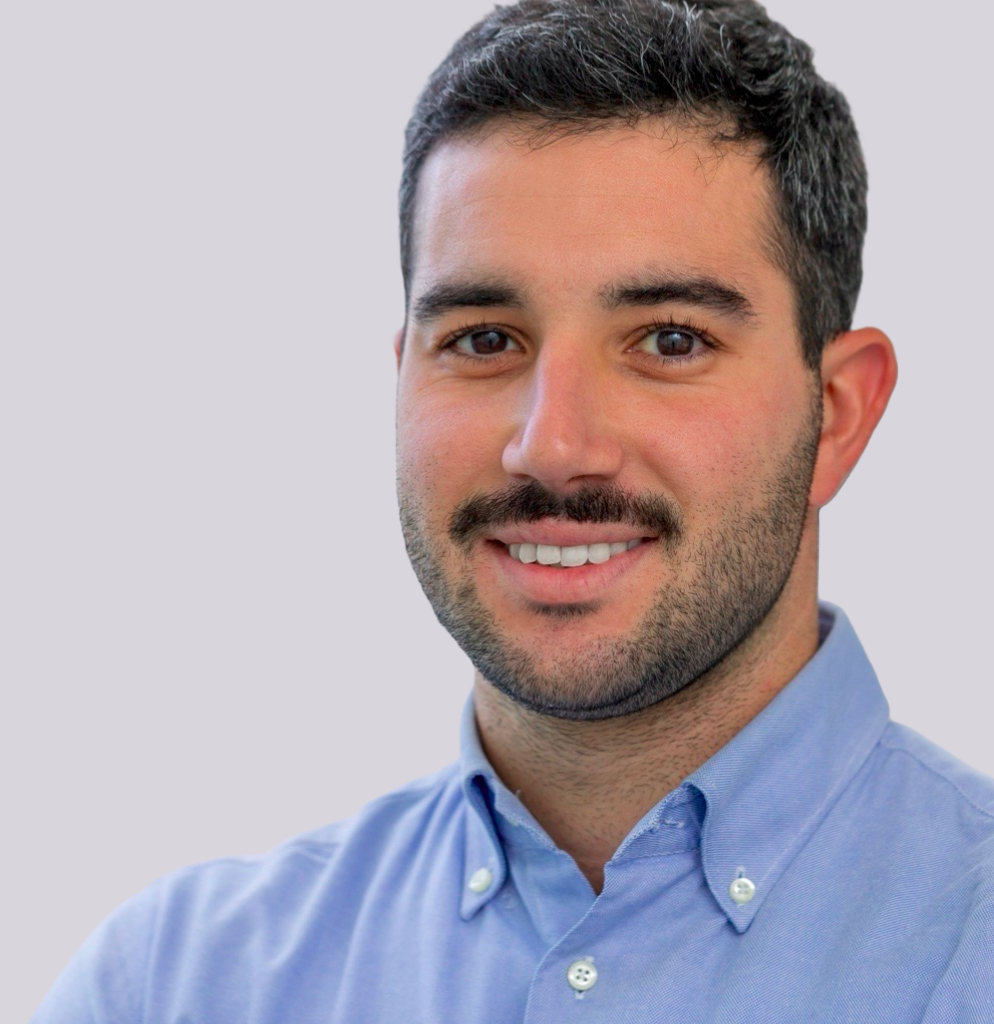}}]{Edoardo Lamon} 
(Member, IEEE) received the Ph.D. degree in robotics from the University of Pisa, Pisa, Italy, in 2021. He is currently an Assistant Professor with the Department of Information Engineering and Computer Science, University of Trento, Trento, Italy, where he is the leading researcher of the Assistive and Medical Robotics group of the \href{https://idra-lab.github.io/}{IDRA Labs}. Until 2023, he acted as a postdoc with the Human–Robot Interfaces and Interaction (HRI$^2$) group, IIT, Genoa, Italy. His interests lie in the intersection of human-robot collaboration, artificial intelligence, robot control, and learning, with the aim of boosting robotics in human-populated environments in medical and industrial settings. Dr. Lamon was awarded as a finalist for the European best thesis in robotics at the Georges Giralt Ph.D. Award 2022 and as a finalist for the best paper award on mobile manipulation at IROS 2022. He serves as Associate Editor for the IEEE Robotics and Automation Letters and for International Conference on Ubiquitous Robots.
\end{IEEEbiography}

\vskip -2\baselineskip plus -1fil

\begin{IEEEbiography}[{\includegraphics[width=1in,height=1.25in,clip,keepaspectratio]{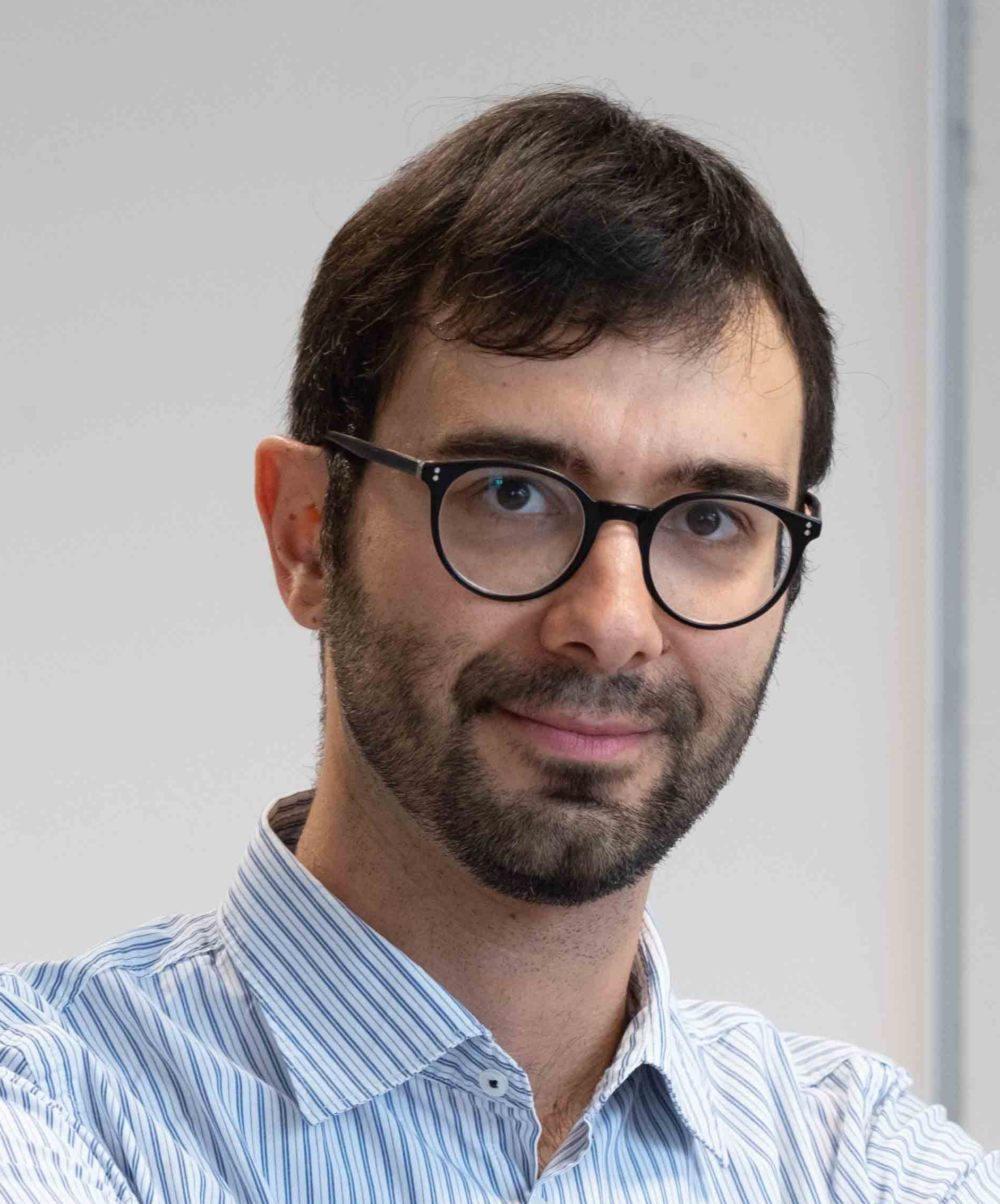}}]{Giacomo Moretti} holds an MSc degree in Energy Engineering, University of Pisa (2013) and a PhD in Mechanical Engineering, Scuola Sant'Anna, Pisa, Italy (2017). He has been a visiting scholar at the University of Edinburgh, UK (2016), and a research fellow at Scuola Sant'Anna (2017-2020) and at Saarland University, Germany (2020-2022) with a Marie-Curie fellowship. Currently, he is an Assistant Professor (with tenure track) in mechanics of machines at the University of Trento, Italy, where he leads a research group on multifunctional material machines. His research interests cover the fields of multifunctional materials (especially electroactive polymers), energy harvesting, and soft robotics.
\end{IEEEbiography}

\vskip -2\baselineskip plus -1fil

\begin{IEEEbiography}[{\includegraphics[width=1in,height=1.25in,clip,keepaspectratio]{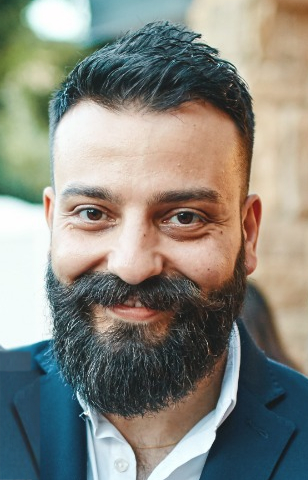}}]{Matteo Saveriano} (Senior Member, IEEE) is an Associate Professor of Control Engineering at the Department of Industrial Engineering of the University of Trento, Italy and the leading researcher of the Intelligent Robotics group of the \href{https://idra-lab.github.io/}{IDRA Labs}. He received the M.Sc. from the University of Naples "Federico II" in 2011 respectively, and a Ph.D. from the Technical University of Munich in 2017. After his Ph.D., he was a post-doctoral researcher at the German Aerospace Center (DLR) and a tenure-track assistant professor at the Department of Computer Science and at the Digital Science Centre of the University of Innsbruck. His research is at the intersection between learning and control and attempts to integrate cognitive robots into smart factories and social environments through the embodiment of AI solutions, inspired by human behaviour, into robotic devices. %He has co-authored more than 70 scientific papers in international journals and conferences. 
He serves regularly as Associate Editor for the main robotics conferences (ICRA, IROS, and HUMANOIDS) and for the IEEE Robotics and Automation Letters, the IEEE Transactions on Robotics, and for The International Journal of Robotics Research. He is the coordinator of the HE EU project \href{https://www.inverse-project.org/}{INVERSE} (GA 101136067).
\end{IEEEbiography}

\vskip -2\baselineskip plus -1fil

\begin{IEEEbiography}[{\includegraphics[width=1in,height=1.25in,clip,keepaspectratio]{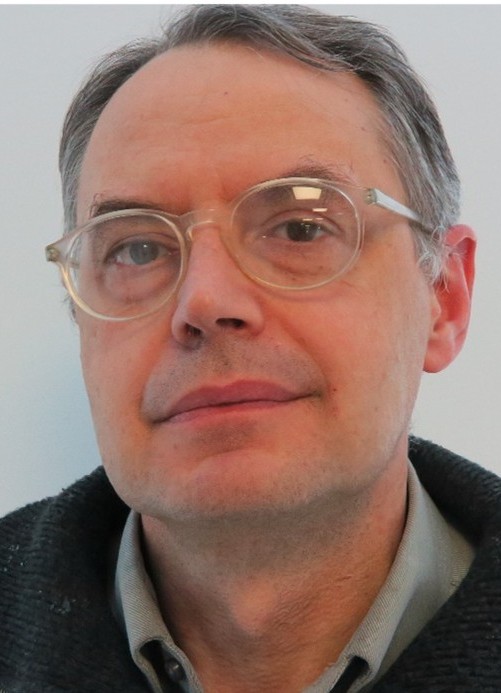}}]{Luca Fambri} 
received the chemistry degree from the University of Parma, Parma, Italy, in 1985.
He is currently an Associate Professor with the Department of Industrial Engineering, University of Trento, Trento, Italy, where he conducts research in polymer science and materials engineering. %In 1993, he was a Visiting Research Associate at Virginia Commonwealth University, USA.
His research interests include polymer processing, additive manufacturing, fibre-reinforced composites, thermal energy storage materials, and biodegradable and recycled polymers, with applications spanning energy, sustainability, and biomedical fields.
Prof. % Fambri is the author or coauthor of more than 140 articles in peer-reviewed international journals, 16 book chapters, and 3 Italian patents and has contributed to over 200 scientific conferences.
He is a member of the National Interuniversity Consortium for Materials Science and Technology (INSTM), the Italian Association for Materials Engineering (AIMAT), and the Italian Association of Macromolecules (AIM).
\end{IEEEbiography}

\vskip -2\baselineskip plus -1fil

\begin{IEEEbiography}[{\includegraphics[width=1in,height=1.25in,clip,keepaspectratio]{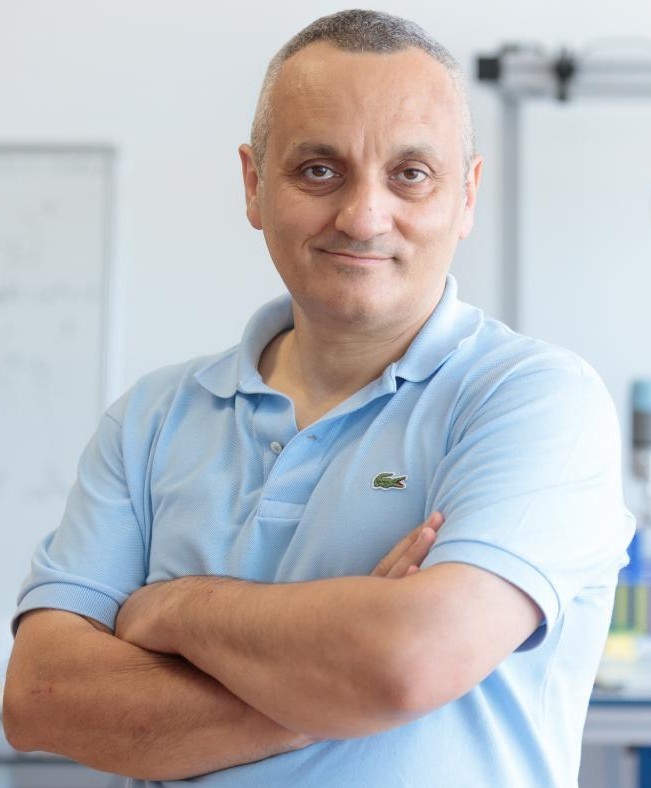}}]{Luigi Palopoli} 
 (Senior Member, IEEE) received the Ph.D. degree in computer engineering from the Scuola Superiore Sant’Anna, Pisa, Italy, in 2002. He is a Full Professor and Director of the Department of Information Engineering and Computer Science, University of Trento, Trento, Italy. %, where he is currently Dean of the Master course in artificial intelligence systems. 
 He has led several industrial and academic research projects, including H2020 ACANTO and FP7 DALi.
 He is the co-founder of \href{https://www.polytecintralogistics.com/en/}{Polytec Intralogistics Srl} and of the \href{https://idra-lab.github.io/}{IDRA Labs}). His main research interests include robotics (with a particular focus on assistive robotics) and embedded system design (with a particular focus on language solutions and probabilistic techniques for soft real–time systems). Prof. Palopoli is currently Associate Editor for IEEE Transactions on Automatic Control and Elsevier Journal of System Architecture. He has also served on the programme committee of different conferences in the area of real-time and control systems.
\end{IEEEbiography}

\vskip -2\baselineskip plus -1fil

\begin{IEEEbiography}[{\includegraphics[width=1in,height=1.25in,clip,keepaspectratio]{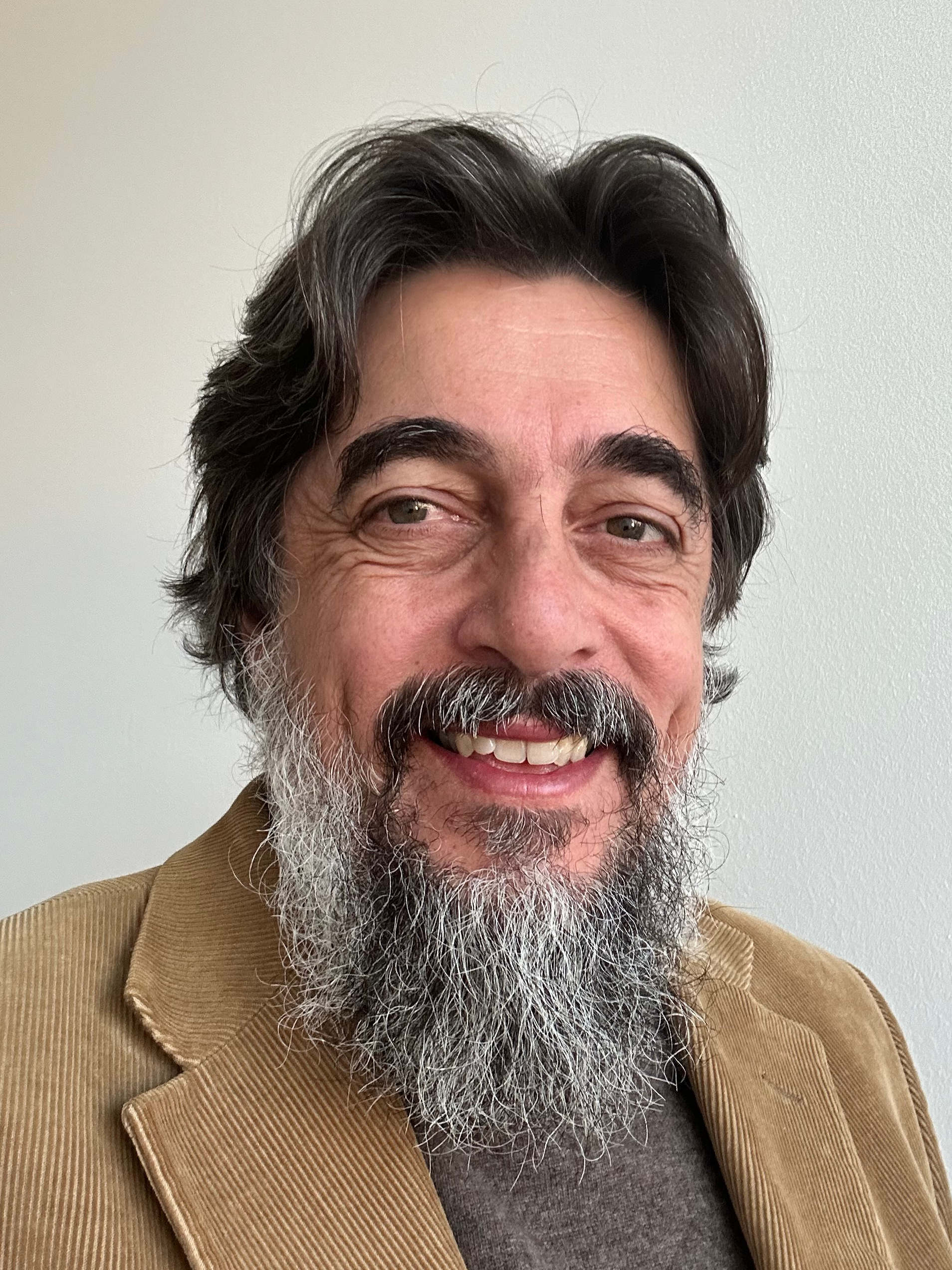}}]{Daniele Fontanelli} 
(Fellow Member, IEEE) received the M.S. degree in Information Engineering in 2001, and the Ph.D. degree in Automation, Robotics and Bioengineering in 2006, both from the University of Pisa, Pisa, Italy.  %He was a Visiting Scientist with the Vision Lab of the University of California at Los Angeles, Los Angeles, US, from 2006 to 2007.  % From 2007 to 2008, he has been an Associate
  % Researcher with the Interdepartmental Research Center
  % ``E. Piaggio'', University of Pisa.
%From 2008 to 2013 he joined as an Associate Researcher the Department of Information Engineering and Computer Science and from 2014 the Department of Industrial Engineering, both at the University of Trento, Trento, Italy, where he is now a Full Professor in the field of measurement and robotics.
From 2008 he joined the University of Trento, Trento, Italy, where he is now a Full Professor in the field of measurement and robotics.
%He has authored and co-authored more than 200 scientific papers in peer-reviewed top journals and conference proceedings.
He is currently a Senior Area Editor for the IEEE Transactions on Instrumentation and Measurement, an Associate Editor for the IEEE Robotics and Automation Letters, and he is a member of the IMEKO TC17 - Measurement in Robotics. He has also served in the technical program committee of numerous conferences in the area of measurements and robotics, and as an Associate Editor for the IET Science, Measurement \& Technology Journal from 2019 to 2024. 
He is the co-founder of \href{https://www.polytecintralogistics.com/en/}{Polytec Intralogistics Srl} and of the \href{https://idra-lab.github.io/}{IDRA Labs}). He is the PI of the EU project \href{https://www.magician-project.eu/}{MAGICIAN} %- iMmersive leArninG for ImperfeCtion detectIon and repAir through human-robot interactioN –
and he was the co-founder and the PI of the EIT-Digital international Master on Autonomous Systems from 2017 to 2023. His research interests include distributed and real-time estimation and control, localisation algorithms, synchrophasor estimation, clock synchronisation algorithms, resource-aware control, wheeled mobile robots, service robotics and human-robot interaction and estimation.
\end{IEEEbiography}

% \end{IEEEbiography}


%\newpage



%\vfill


\end{document}